\documentclass[runningheads]{llncs}

\usepackage{eccv}

\usepackage[toc,page,header,titletoc]{appendix}
\usepackage{multicol}
\usepackage{multirow}
\usepackage{subcaption} 

\usepackage{eccvabbrv}

\usepackage{graphicx}
\usepackage{booktabs}

\usepackage{graphicx}
\usepackage{amsmath}
\usepackage{bbm}
\usepackage{amssymb}
\usepackage{booktabs}
\usepackage{color}
\usepackage{nicefrac}

\usepackage{etoolbox}
\makeatletter
\pretocmd{\chapter}{\addtocontents{toc}{\protect\addvspace{15\p@}}}{}{}
\pretocmd{\section}{\addtocontents{toc}{\protect\addvspace{5\p@}}}{}{}
\pretocmd{\subsection}{\addtocontents{toc}{\protect\addvspace{3\p@}}}{}{}
\pretocmd{\subsubsection}{\addtocontents{toc}{\protect\addvspace{2\p@}}}{}{}
\makeatother
\usepackage{titletoc}
\newcommand{\algname}{{{GUMP}}\xspace}

\usepackage{colortbl} 
\usepackage{multirow} 
\usepackage{resizegather} 
\definecolor{Gray}{gray}{0.85} 

\usepackage{amssymb}
\usepackage{pifont}
\newcommand{\cmark}{\ding{51}}
\newcommand{\xmark}{\ding{55}}
\usepackage{xcolor}

\usepackage{caption}
\usepackage[accsupp]{axessibility}  

\usepackage[pagebackref,breaklinks,colorlinks]{hyperref}

\usepackage{orcidlink}

\usepackage{hyperref}

\begin{document}

\title{Solving Motion Planning Tasks with a Scalable Generative Model}

\titlerunning{GUMP}

\author{Yihan Hu\ \and
Siqi Chai\and
Zhening Yang \and Jingyu Qian \and Kun Li \and Wenxin Shao \and Haichao Zhang \and Wei Xu \and Qiang Liu}

\authorrunning{Y. Hu et al.}

\institute{Horizon Robotics Inc. \\
\email{\{yihan.hu96, Proliu\}@gmail.com}}

\maketitle

\begin{abstract}
As autonomous driving systems being deployed to millions of vehicles, there is a pressing need of improving the system's scalability, safety and reducing the engineering cost. A realistic, scalable, and practical simulator of the driving world is highly desired. In this paper, we present an efficient solution based on generative models which learns the dynamics of the driving scenes. With this model, we can not only simulate the diverse futures of a given driving scenario but also generate a variety of driving scenarios conditioned on various prompts. 
Our innovative design allows the model to operate in both full-Autoregressive and partial-Autoregressive modes, significantly improving inference and training speed without sacrificing generative capability. This efficiency makes it ideal for being used as an online reactive environment for reinforcement learning, an evaluator for planning policies, and a high-fidelity simulator for testing. We evaluated our model against two real-world datasets: the Waymo motion dataset and the nuPlan dataset. On the simulation realism and scene generation benchmark, our model achieves the state-of-the-art performance. And in the planning benchmarks, our planner outperforms the prior arts. We conclude that the proposed generative model may serve as a foundation for a variety of motion planning tasks, including data generation, simulation, planning, and online training.  Source code is public at \url{https://github.com/HorizonRobotics/GUMP/}.

\end{abstract}

\section{Introduction}

Autonomous driving (AD) has evolved from a visionary concept to tangible products in recent years~\cite{mobileye2024ces, tesla2022aiday,xpeng2023CVPR}. 
Despite these advancements, challenges in technical scalability remain~\cite{hawke2021reimagining}. These involve the system's ability to adapt to new and unseen environments, which is the root cause to ongoing safety concerns and the extensive engineering efforts required to address various failure scenarios.
 To continuously scale up in terms of safety and cost efficiency, developing a model that learns and represents the driving world is essential. Such a model could enable the continuous generation of data for training and testing, moving toward a more integrated and learned driving system that seamlessly interacts with real-world road users.

A driving world may be considered as a sequence of driving scenarios, which typically contain a map and multiple agents interacting within that space. Obtaining a local map for these scenarios is generally straightforward, whether through map providers~\cite{OpenStreetMap} or through online mapping modules~\cite{MapTR, maptrv2, liu2022vectormapnet}. The real challenge, however, lies in understanding the dynamic nature of traffic - specifically, the interdependence among road users. Thus, a fundamental task in modeling the driving world involves accurately predicting the behaviors of these agents, formulating the plans based on these predictions and continuously refining these predictions as the agents take action.

This task has been intensively explored in recent years, including motion forecasting~\cite{gu2021densetnt, ngiam2021scene, Nayakanti23icra_Wayformer, shi2022motion, hu2023_uniad} and imitative traffic modeling~\cite{suo2021trafficsim, li2023drivingdiffusion}. All of these works are able to model the either marginal or joint distributions of the multi-agents' future trajectories. Despite these advancements, these models are typically trained in an open-loop setting, which may limit their ability to adapt to out-of-distribution states encountered in real-world, closed-loop settings. Consequently, their capability for traffic simulation is generally constrained to short duration, such as a few seconds.

Recently, closed-loop motion prediction has attracted increasing attentions, which addresses the problems of long horizon traffic simulation and interactive simulation~\cite{montali2023waymo,Wang23tr_JointMultipathSimAgents,seff2023motionlm,philion2023trajeglish}. Through the iterative forecasting of each agent's next states, autoregressive (AR) models articulate the interactions among agents as a series of cascading conditional distributions. At each subsequent time step, sampling from the model can generate a variety of actions for an agent, aligning with both the context of the scene and the preceding actions of other agents. Therefore, learning the dynamics of the driving world is approached in the same manner as learning sequence prediction, akin to the techniques used in language modeling~\cite{radford2019language}.

In light of these work, we propose the \textbf{G}enerative \textbf{U}nified model for \textbf{M}otion \textbf{P}laning tasks, namely \algname, with a generative model structure and a simple tokenizer, which uses an object's unique ID as its key and a compressed state space as its value. This design significantly enhances the model's flexibility, enabling us to adopt an efficient partial AR structure. This model demonstrates high scalability in computational efficiency, and shows strong generalization capabilities in understanding complex traffic flows and handling long-tail cases. In addition, its generative ability allows infinite sampling from the learned distribution, which enables long-horizon reactive agent simulation.

\begin{figure}
    \centering
    \vspace{-5mm}
                          \includegraphics[width=0.35\linewidth]{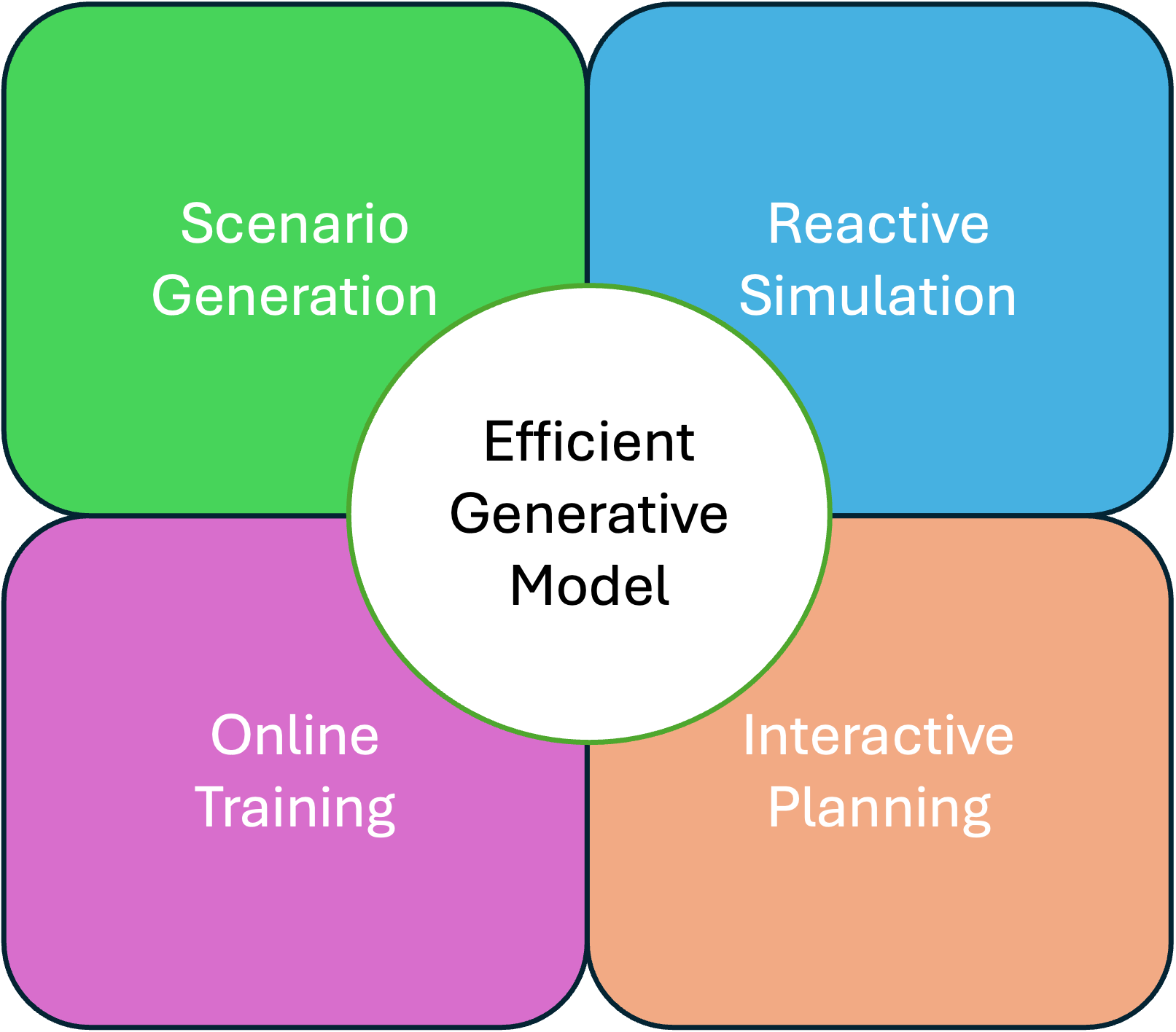}
    \caption{We are motivated to provide a generative model as the central unit that supports all the learning-based motion planning tasks in the autonomous driving domain. We categorize the tasks into four distinct sub-domains: data generation, model evaluation, model training, and model inference. These sub-domains are visually distinguished in our diagram by different colors—green for data generation, blue for model evaluation, purple for model training, and orange for model inference. Our approach encompasses both offboard applications (the first three sub-domains) and onboard application (the last sub-domain). Specifically, scene generation aims at data generation capable of producing specific traffic scenarios based on context information, such as high-definition maps or user prompts; Reactive simulation aims at a closed-loop evaluator that provides realistic, human-like agents that respond to the behavior of the ego vehicle and its environment; Online training aims at a closed-loop training module that allows the learned policy to interact with environment, collect rewards, and perform back-propagation.
    Lastly, interactive planning aims at enhancing an onboard planner by parallel unrolling to seek for the optimal trajectory that achieves the highest reward.}
    \label{fig:eccv2024_fig1}
\vspace{-12mm}
\end{figure}

Centered at this model, as shown in Fig.~\ref{fig:eccv2024_fig1}, we have explored various downstream tasks, and found that this model can be used as:
\begin{enumerate}
\item a \emph{data generator} that creates scenarios specific to user's prompts;
\item a \emph{realistic simulator} serving as a reactive closed-loop test bed;
\item a \emph{planner} that unravels interactions between agents to reduce infractions and improve human-like behavior;
\item an \emph{online training} module that enhances the effectiveness of reinforcement learning for policy models.
\end{enumerate}

In summary, our main contributions are threefold:
Firstly, we propose a novel generative model that features a simple key-value paired tokenizer. We show that this model achieves state-of-the-art performance on both the simulation and planning benchmarks.
Secondly, we have extensively investigated the usage of this model as a foundation model in a wide range of downstream tasks, and demonstrate that it can significantly improve the functionalities of these tasks.
Thirdly, we provide a framework of using the generative model as a central component for developing a closed-loop training and evaluation system.
To our best knowledge, we are the first to solve all learning-based motion planning tasks with a unified framework.

\section{Methods}
\subsection{Formulation}
Let the dynamic states of the agents at time $t$ be denoted by $s_t = (a_t^{AV}, a_t^{env})$, where $a_t^{AV}$ is the self-driving vehicle, and $a_t^{env}$ is composed of surrounding agents. Specifically, $a_t^{AV} = (a_t^0)$ and $a_t^{env} = (a_t^i) \quad \text{for} \quad i \in \{1, 2, \ldots, n\}$, and the context is denoted as $c$, which includes a static map~\cite{OpenStreetMap} and language description prompts~\cite{sima2023drivelm}. We factorize the joint probability distribution of traffic scenarios as follows:

\begin{multline}
    P(s_t, s_{t-1}, \ldots, s_0, c) = \\
    \underbrace{P(s_t | s_{t-1}, \ldots, s_0, c) \cdot \ldots \cdot P(s_1 | s_0, c)}_{\text{scene extrapolation}} 
    \cdot \underbrace{P(s_0 | c)}_{\text{scene generation}} 
    \cdot \underbrace{P(c)}_{\text{context}}.
\end{multline}

Naturally, traffic scenarios can be modeled in a probabilistic and sequential manner: given just the context $c$, we can generate the initial states of the agents $s_0$; With both the context $c$ and the initial states $s_0$, we can extrapolate the subsequent states ${s_1, s_2,\dots, s_{t-1}}$ of the agents. The initial step allows us to create scenarios through scene generation, while the next step enables us to foresee how these scenarios evolve over time, which we refer to as scene extrapolation.

\subsubsection{Scene Generation}
The initial states can be further factorized as:

\begin{equation}
     P(s_0 | c) = \prod_{j} P(a_0^j | a_0^0, \ldots, a_0^{j-1}, c) .
\end{equation}

Specifically, we autoregressively generate the initial state of each agent in the scene, denoted as \(a_0^i\), where \(a_0^i = \{x, y, \theta, v_x, v_y, w, l\}\) represents the initial position (\(x, y\)), heading (\(\theta\)), velocity components (\(v_x, v_y\)), width (\(w\)), and length (\(l\)) for the \(i^{th}\) agent.

\subsubsection{Scene Extrapolation} We can further factorize the the scene extrapolation task as ~\cite{montali2023waymo}:

\begin{multline}
    P(s_T, s_{T-1}, \ldots | s_0, c) = P(a_{1:T}^{AV}, a_{1:T}^{envs} | s_0, c) \\
    = \prod_{t} \pi_{AV}(a_t^{AV} | a_{<t}^{AV}, a_{<t}^{env}, c) \cdot P(a_t^{envs} | a_t^{AV}, a_{<t}^{AV}, a_{<t}^{env}, c).
    \label{eq:simagents}
\end{multline}

In this model, \(\pi_{AV}\) represents the policy for autonomous vehicles. Note that this policy can be replaced with any planning policy. Moreover, we can categorize the environmental actions \(a_{t}^{env}\) into two distinct groups: \(a_{t}^{tracked}\) and \(a_t^{newborn}\). Here, \(a_{t}^{tracked}\) refers to objects that have been previously tracked, while \(a_t^{newborn}\) represents objects that have just emerged in the environment.

We enable the probabilistic modeling of dynamic scene information without imposing a limit on the number of objects throughout the extrapolation. This capability is crucial for realistically modeling scenarios where objects may disappear or newly appear, addressing both currently obscured objects that may become visible later and those present now but might move out of view. Such a nuanced treatment of dynamic elements significantly improves our model’s capability to handle the complex, unpredictable, and long-tail problems in autonomous driving.

\subsubsection{Evaluate Planning Policies} 
\label{sec: formulation planning}

Suppose we have a series of candidate policies, denoted as \(\pi_{i}\), where \(i = \{ 0,1,\dots, N_{p} - 1\}\). These policies can be rule-based or learning-based. Given the observations  \(s\) at a certain moment, these policies can output different response actions \(a_i \sim \pi_{i}(s)\). Under the Markov assumption, the value function $V^{\pi_i}(s)$ of a chosen policy $\pi_i$ is denoted as:

\begin{equation}
\label{equ: value}
    V^{\pi_i}(s) = \sum_{s'} P(s'|s,\pi_i(s)) \left[ r(s, \pi_i(s)) + \gamma V^{\pi_i}(s') \right]
\end{equation}
where $\gamma$ is the discount factor and \(P(s'|s,\pi_i(s))\) is the state transition probability function, which can be modeled with the probabilistic rollouts of our world model. Here, \(s'\) represents the next state, and \(r(s, \pi_i(s))\) is the reward function, which may be defined as \cite{caesar2022nuplan}:

\begin{equation}
\label{equ: reward}
 r(s, \pi_{i}(s)) = \prod_{m} \Theta_{m}(s, \pi_{i}(s)) \cdot \sum_{n} \omega_{n} \Phi_{n}(s, \pi_{i}(s))
\end{equation}
where $\Theta$ is the critical metrics, such as safety and driving directions, and $\Phi$ is less critical weighted metrics, such as progress, speed limit, comfort and so on, and $\omega_{n}$ is the metric weight. More details can be found in the Appendix~\ref{sec:nuplan_metrics}.

Given the vast and probabilistic nature of the state space \(s\), we estimate it by sampling from its overall distribution. As in  Eq.~\ref{eq:simagents}, the results are referred as rollouts, denoted as $X_i^{k}=\{s_0, s_1, \ldots, s_T\}_i^{k}=\{a_t^{AV},a_t^{env}\}_i^{k}$, where $i \in \{0, 1, 2, \ldots, N_{p} - 1\}$, and $k \in \{0,1,2,\ldots, N_{r}-1\}
$, and $N_{p}$ is the number of the candidate policies, and $N_{r}$ is the number of the paralleled rollouts for each policy. So, the total number of the rollouts is $N_{p} \times N_{r}$ . Then we apply the evaluator to calculate the accumulated rewards for each policy following Eq.~\ref{equ: value}.

Accordingly, the ego vehicle's optimal driving policy can be written as
\begin{equation}
    \pi^{\star}(s_0) = \underset{\pi_i}{\mathrm{argmax}} \, \hat{V}^{\pi_i}(s_0)
\end{equation}
where $\hat{V}^{\pi_i}(s_0)$ is the value estimated through rollouts obtained by sampling.

\subsubsection{Online Training with Reinforcement Learning}
\label{sec: formulation RL}
In standard Reinforcement Learning (RL), the problem to be solved is typically described as an Markov Decision Process (MDP)~\cite{sutton_book}, which is commonly defined by a tuple $(\mathcal{S, A}, P, r, \gamma)$, where $\mathcal{S}$, $\mathcal{A}$ denote the state space and action space respectively. 
$P(s'|s, a)$ denotes the state transition process, \emph{i.e.}, the probability of transiting to
$s'$ given the current state $s$ and action $a$. $r$ is the reward function and $\gamma$ is the discount factor.
The goal of RL is to learn a policy $\pi(s;\theta)$ that maximizes the accumulated discounted return.

While much effort has been devoted to the development of RL algorithms in the past~\cite{schulman2017proximal, haarnoja2018soft}, the importance of high-fidelity simulator starts to gain more attentions recently~\cite{when_to_trust_simulator}.
For autonomous driving,
one challenge in terms of simulation is realistic reactive behaviors
from other road participants (\emph{i.e.} the implementation of $P$), which is arguably hard to be fully characterized by scripted rules~\cite{dosovitskiy2017carla}. Therefore, instead of implementing $P$ by scripting (\emph{i.e.} $P\!\triangleq\! P_{\rm scripted}$) as in standard simulator~\cite{dosovitskiy2017carla}, 
we leverage our data-driven world model for the state transition:
\begin{equation}
P(s'; s, a) \triangleq P_{\rm WorldModel}(s'; s, a),
\end{equation}
where $P_{\rm WorldModel}$ denotes our world model trained on real-world data, therefore capturing the real-world behavior and avoiding cumbersome rule crafting. This could significantly enhance RL by ensuring alignment with real-world logic.

The unroll process could be presented as the following equations:
\begin{equation*}
\begin{array}{ll}
s' \sim P_{\rm WorldModel}(s'; s, a) \qquad \qquad  & a \sim \pi(s; \theta).
\end{array}
\end{equation*}
And the rewards can be calculated with Eq.~\ref{equ: value} and Eq.~\ref{equ: reward}.
For training, one may use typical RL algorithms such as  Soft Actor-Critic (SAC)~\cite{haarnoja2018soft}.

\subsection{Network}

\begin{figure*}[h]
\begin{center}
\vspace{-8mm}
\includegraphics[width=0.65\textwidth]{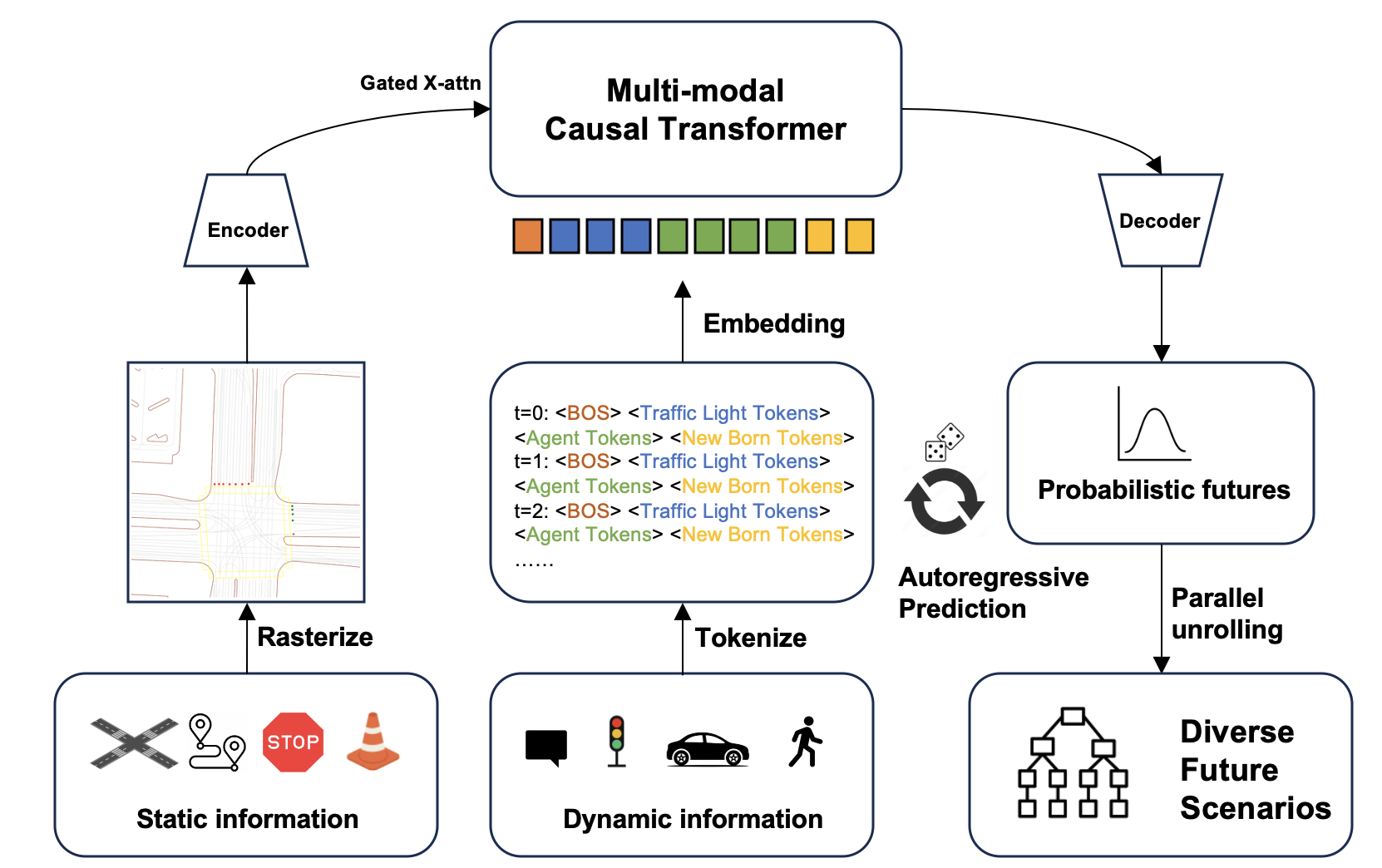}
\end{center}
\caption{\algname is composed of a raster encoder that encodes static information including map, route and static objects, a key-value pair tokenizer that discretizes dynamic information including the states of road users and traffic lights, a Multimodal Causal Transformer (MCT) that predicts the next latent embedding based on key queries in an autoregressive manner, and a decoder that samples the probabilistic features and decodes to future scenarios.}
\vspace{-6mm}
\label{fig:architecture}
\end{figure*}

The overall architecture of our model, depicted in Figure~\ref{fig:architecture}, is comprised of four main components: a static raster autoencoder, a dynamic tokenizer, a Multimodal Causal Transformer (MCT), and an auto-regressive decoder. For the static information, elements such as maps, navigation routes, and static obstacles are converted into a raster image. This image is then encoded with a 2D Convolutional Encoder~\cite{he2016deep}. Meanwhile, dynamic information is encapsulated into a sequence of tokens, providing a linguistic representation of driving behaviors. The static and dynamic information are then fused~\cite{alayrac2022flamingo} and encoded with a multimodal visual-language model~\cite{radford2019language}. Due to its predictive nature, our model can be further accelerated by the intra-frame non-autoregressive (NAR) conversion.

\subsubsection{Tokenization}
In probabilistic modeling of continuous state spaces, discretization is required. Unlike previous studies that use complex transformations to map the state space to the action space and then reintegrate to recover the state space~\cite{seff2023motionlm, philion2023trajeglish}, our method directly quantizes the state space.

Choosing state space over action space is motivated by several key factors. Although the action token space is more compact, featuring a smaller vocabulary size, this compactness often sacrifices interpretability. Moreover, encoding with action space requires state-dependent decoding, where each step depends on all preceding actions and the initial states. This dependency may result in compounded errors. Additionally, this dependency can also limit the network's flexibility, thereby restricting its generative capabilities, such as predicting traffic lights, handling the emergence and disappearance of elements, or generating scenarios.

In our design, each object is composed of two tokens with distinct functionalities, akin to a ``key-value pair'': a control token and a state token. The control token serves as the key, used to distinguish between different objects by summing the token embeddings of each object's unique ID and object category. The state token serves as the value, utilized to depict the specific state space of the object. Specifically, we select different state spaces for different types of objects, and perform quantization of the state space for each object. For instance, for traffic lights, the token space includes coordinates ($x, y, \theta$) and traffic signal states ($s_{tl}$). For traffic participants like pedestrians and vehicles, the token space encompasses coordinates ($x, y, \theta$), velocity ($v_x$, $v_y$), and size ($w$, $l$). This key-value pair tokenization enables our model with the indexing ability. We can selectively query and decode any object of interest or arbitrarily add or remove objects. This approach allows the model to dynamically allocate computational resources for increased efficiency and to manage the disappearance and emergence of objects effectively.

Similar to language models in sequence modeling, we incorporate special tokens, including the beginning of sequence (BOS) token, traffic light end token, and newborn begin token. The final tokenized sequence structure can be depicted as shown in Fig.~\ref{fig:architecture}. Additionally, we use a predetermined set of scenario description prompts, which can be represented with a fixed vocabulary. This vocabulary can be embedded similarly to the special tokens.

\subsubsection{Token Embedding}
\label{sec: embedding}
Each token is embedded into latent features. For the key token of each agent, we sum the embedding features of the agent's unique ID and its class type. For it's value token, we embed all the states with sinusoidal embedding along with tokenized state embeddings. All embeddings are then summed with a learnable positional embedding, as in the equation below:
\begin{equation}
\begin{aligned}
e_{\text{key}}^i &=  \text{PE}(i) + \sum_{s \in \{\text{id}, \text{class}\}}\text{Embed}(s) \\
e_{\text{value}}^i &=  \text{PE}(i) + \sum_{s \in \{x, y, \theta, v_x, v_y, w, l\}}(\text{Embed}(s) + \text{sinusoidal}(s))
\end{aligned}
\end{equation}

 Here, $i$ is the position index in the token sequence, PE is the learnable positional embedding, $sinusoidal$ is the sinusoidal embedding and $Embed$ is the learnable token embedding. For static objects such as traffic light, we only embed the coordinates ($x$, $y$, $\theta$) and the traffic light status ($s_{tl}$).

\subsubsection{Multi-modal Causal Transformer}
The Multi-modal Causal Transformer (MCT) module forms the generative core. It builds on the GPT-2 architecture~\cite{radford2019language,karpathy_nanoGPT} and includes Gated Cross Attention (GCA) Blocks~\cite{alayrac2022flamingo} for fusing information from different modalities. We intermittently insert the GCA blocks among self-attention layers to progressively and interactively fuse the dynamic and static information.

\subsubsection{RNN Token Decoder}
To decode the discrete state space \(S\) from the compact latent embedding produced by the Multi-modal Causal Transformer (MCT), an additional GRU decoder is utilized after the MCT. By querying the MCT with the key embedding for each agent, we extract the corresponding state latent space, which is then autoregressively decoded using stacked GRU layers according to the sequence \(\{ w, l, v_x, v_y, x_0, y_0, \theta_0, ..., x_T, y_T, \theta_T\}\), where \(T\) represents the predicted horizon for each agent. This process facilitates temporal aggregation and intra-frame NAR conversion. To maintain the stability of the decoding, we adopted temperature-scaled top-k sampling strategy~\cite{Hinton2015DistillingTK, fan2018hierarchical} and dynamic valid masking to ignore invalid tokens.

\subsubsection{Prediction Chunking and Temporal Aggregation}
\begin{figure}[h]
\vspace{-5mm}
\centering
\includegraphics[width=0.45\textwidth]{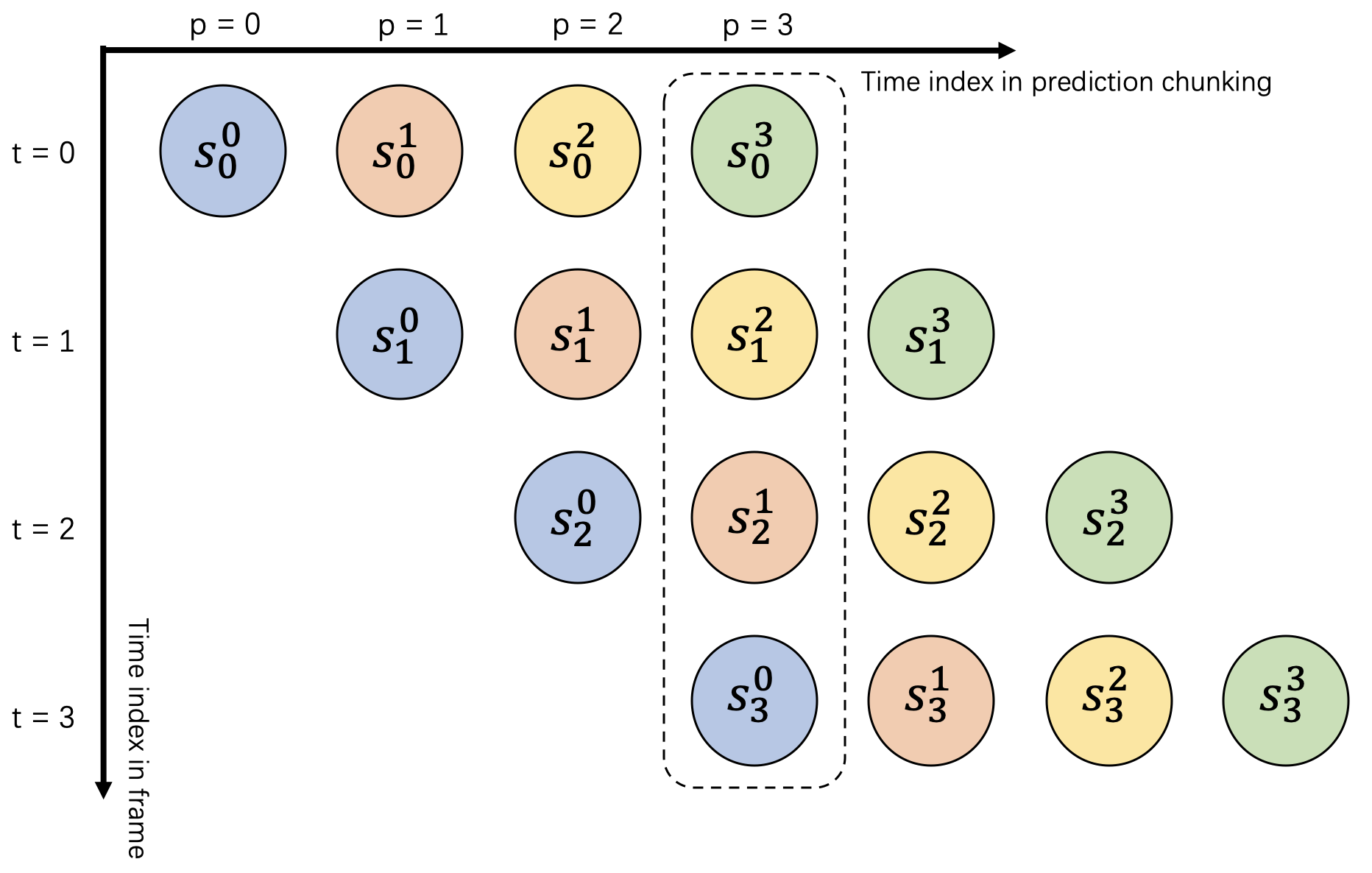}
\caption{Prediction Chunking and Temporal Aggregation. \(s_t^{p}\) denotes the per-agent state at time step \(t\) and time index \(p\) within the chunking data. In our context, \(s_t^{p}\) comprises the set \(\{x, y, \theta\}\).}
\label{fig:chunking}
\vspace{-5mm}
\end{figure}

To combat the compounding errors and stabilize the AR process, inspired by the concept of ``action chunking'' in~\cite{zhao2023learning}, we have developed a prediction chunking and temporal aggregation module. We perform the chunking process by predicting longer time horizon with the light-weight RNN described above. During the AR decoding process, we aggregate all the prediction results of each time step as illustrated in Fig.~\ref{fig:chunking}. Specifically, we take a weighted average across the different time steps with a decay rate $\gamma$; the final output state can be represented as $\hat{s}_t = \sum_{i=0}^{T} \gamma^{i} \cdot s_{t-i}^{i}$, where $t$ is the output time step, and $T$ is the total time steps of the chunking data. This module significantly improved the model's performance on the Waymo Sim Agents metric~\cite{montali2023waymo} as the experimental result shown in Appendix.~\ref{sup:chunk}.

\subsubsection{Intra-frame NAR Conversion}
\label{sec:convert}
\begin{figure}[h]
\begin{center}
\includegraphics[width=0.9\textwidth]{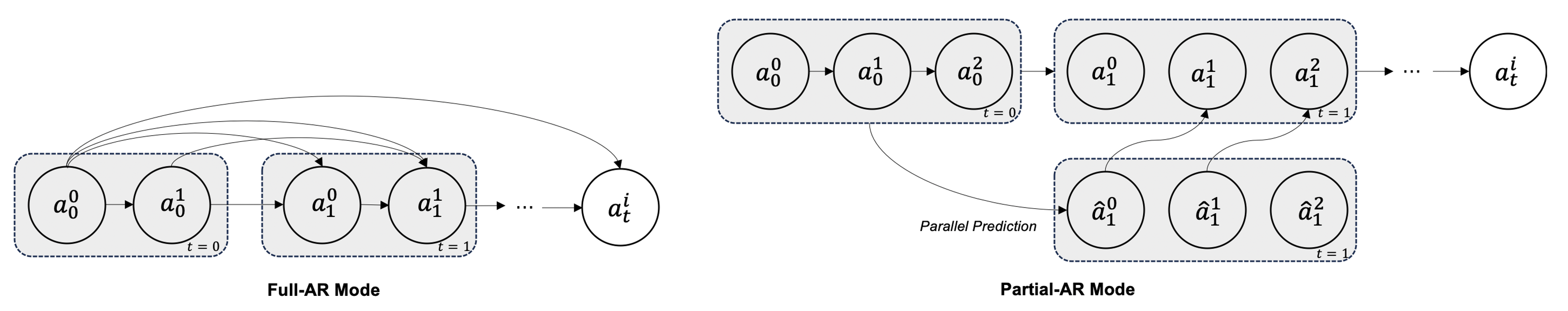}
\end{center}
\caption{This figure compares the full-AR mode with the partial-AR mode. Here, $a_{t}^i$ represents the state of the $i^{th}$ agent at time $t$. Employing a GRU decoder alongside prediction chunking, we can simultaneously predict the next state of each agent, denoted as $\hat{a}_{t+1}^{i}$. These predictions serve as surrogate conditions to bypass intra-frame sequential dependencies, markedly speeding up the process by eliminating the need for an intra-frame sequential AR procedure.}
\label{fig:nar}
\vspace{-5mm}
\end{figure}

To speed up a full-AR model, one approach is to decode the parts that are less dependent in parallel~\cite{chang2022maskgit}. Considering the characteristics of the traffic simulator, we temporarily ignore the current interactions between agents and predict the next state of each agent in parallel with the GRU decoder mentioned earlier. By conditioning on these parallel-decoded states instead, we can eliminate the intra-frame sequential dependency of the AR process, as shown in Fig.\ref{fig:nar}. We can then compensate for the intra-frame dependencies by a single forward pass and solve for more precise agent states efficiently. By converting the full-AR to partial-AR mode, we achieve a significant speedup, as demonstrated in Appendix~\ref{sup: nar ablation}. Moreover, this conversion is optional, allowing our model to maintain its AR ability, which is essential for scene generation tasks and managing newborn agents. Also, our method avoids altering the causal attention mask, therefore is highly compatible with acceleration libraries~\cite{dao2022flashattention,dao2023flashattention2}, leading to notable improvements in training and inference speed and memory usage compared to the versions implemented with customized masking strategy.

\subsection{Framework}
\begin{figure}[h]
\begin{center}
\includegraphics[width=0.8\textwidth]{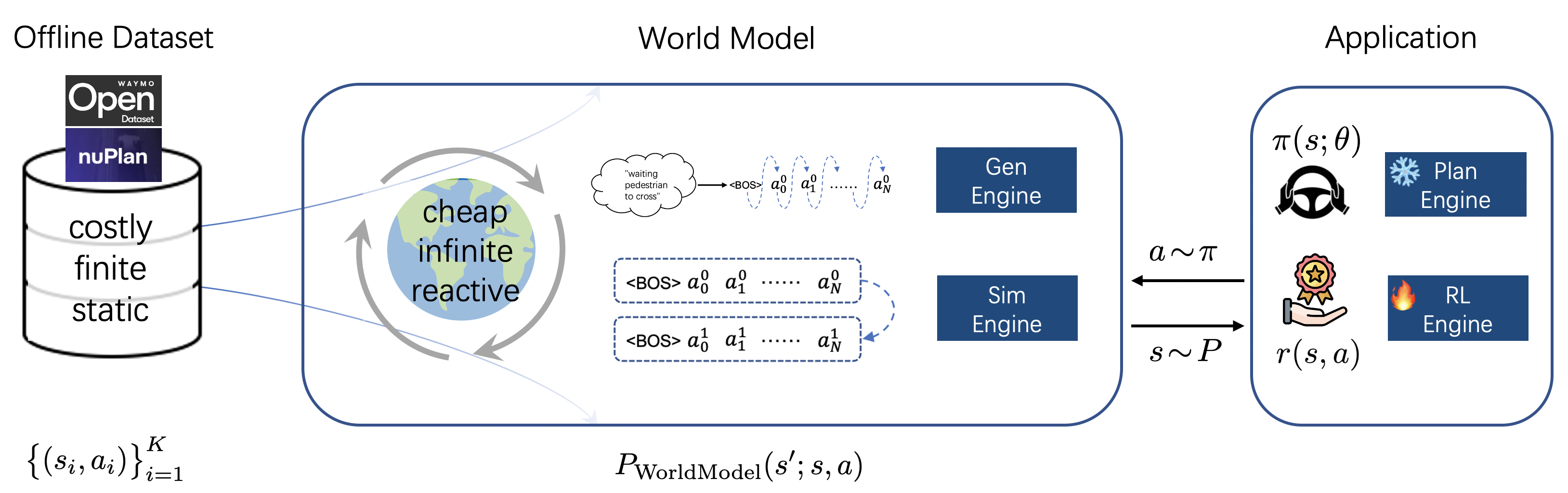}
\end{center}
\caption{\algname serves as a central unit, bridging offline datasets with downstream applications. By learning from the collected offline data, we utilize a generative framework to produce a vast amount of affordable, interactive data, which benefits various downstream tasks such as scene generation, reactive simulation, planning, and online training.}
\label{fig:framework}
\vspace{-8mm}
\end{figure}

In this section, we introduce how \algname serves as a central unit to support a series of downstream tasks, as illustrated in Fig.~\ref{fig:framework}. For those tasks, we have developed specific engines to interact with \algname, categorized into scene generation, scene extrapolation, planning, and RL engines.

For scene generation, \algname incorporates a scenario prompt as an additional condition, as described in Sec.~\ref{sec: embedding}, operating in full-AR mode.  Specifically, we sequentially construct queried key tokens based on the unique ID and the category of each agent. By querying the MCT, we decode the latent features into initial states.  At each iteration, we query and decode information for a single object and use it as the condition for the next query, repeatedly. This AR generation approach effectively manages the 
 relationships between the dynamic objects and the context, and can be well controlled by users. 

Different from the task of scene generation, scene extrapolation uses historical scenes as conditions, which can be either from log data or generated by the scene generation task. To accelerate the process, the MCT operates in a partial-AR mode using intra-frame NAR conversion. Through scene extrapolation, we are able to generate diverse, interactive, and closed-loop state predictions for multiple objects within the scene.

This realistic and efficient simulator, denoted as ``Plan Engine'' in Fig.~\ref{fig:framework}, can be used both off-board as a policy evaluation environment and onboard for deployment, further enhancing existing policies to obtain a more powerful planning policy. This enhancement is achieved by overriding the next state of the ego vehicle in the ``Sim Engine'' with the action (i.e ``$a$'' or ``next ego state'') output by the existing driving policy, facilitating interaction between the policy and the world model. The resulting rollouts (i.e ``$s$'' or ``current states'') from this interaction are utilized by the reward module to compute relevant scores. After several rounds of interaction, the model's performance is assessed based on the accumulative score, which can be used for evaluating off-board models or selecting online proposals, as described in Sec~\ref{sec: formulation planning}. This highly realistic and probabilistic simulator accurately simulates environmental changes, enabling more precise evaluation and selection of optimal policies.

Different from the ``Plan Engine'', the ``RL Engine'' employs a trainable ML-policy, which undergoes RL training via accumulative rewards. Unlike previous approaches that used IDM\cite{li2022metadrive} or logsim\cite{chen2021learning}, the simulator powered by \algname offers a more efficient, realistic, diverse, and interactive environment. This reduces the simulation-to-reality domain gap and enhances the model's generalization capability.

\section{Experiments}
Our experiments are primarily conducted on two mainstream public datasets: the Waymo Open Dataset (WOD)~\cite{Ettinger_2021_ICCV,Kan_2023_arxiv} and the nuPlan~\cite{caesar2022nuplan}. Our experiments cover scene generation, trajectory prediction, WOD sim agents, and planning. We primarily use the Waymo dataset for the scene generation~\cite{feng2023trafficgen,suo2021trafficsim,tan2021scenegen}, trajectory prediction and sim agents~\cite{montali2023waymo} tasks, and use the nuPlan dataset for prompts conditioned scene generation and planning. You can find more information about these datasets and our benchmarks in the Appendix~\ref{sec: dataset and benchmark}, and also related experiment settings and training details Appendix~\ref{sec: experimrnt setup}. We have placed more ablation studies, experimental results regards trajectory prediction and qualitative results in Appendix~\ref{sec: ablation study},\ref{sup: trajectory} and \ref{sup: quanlitative},  respectively.

\subsubsection{Scene Generation}
As shown in Table~\ref{tab:mmd}, we compare our model with other methods on the WOD motion dataset. To ensure a fair comparison, we have closely followed the experimental setting of TrafficGen~\cite{feng2023trafficgen}. Our results significantly outperformed all other competitors across all metrics by a large margin, especially in terms of speed and size, where the error was reduced by 42.1\% and 26.6\% respectively compared to the previous state-of-the-art performance. This fully demonstrates our model's exceptional capability in scene generation, capable of creating realistic data.

\begin{table}[h]
\centering
\resizebox{0.8\textwidth}{!}{
\begin{tabular}{l|cccc}\hline
Method & Position$\downarrow$ & Heading$\downarrow$ & Speed$\downarrow$ & Size$\downarrow$\\
\hline
SceneGen\cite{tan2021scenegen} &0.1362 &0.1307 &0.1772 &0.1190\\
TrafficGen\cite{feng2023trafficgen} & 0.1192 &0.1189 & 0.1602 &0.0932\\
$\text{TrafficGen}^{\star}$ & 0.1221 &0.1174 & 0.1661 & 0.0913\\
\textbf{\algname-m} & \textbf{0.1107}(-9.3\%) &  \textbf{0.099}(-15.7\%) & \textbf{0.0961}(-42.1\%) & \textbf{0.0670}(-26.6\%)\\
\hline
\end{tabular}
}
\caption{Maximum mean discrepancy (MMD) results on WOD Motion Dataset: Lower MMD Indicate Better Performance Across All Metrics. $\star$: Results reproduced by us under the same experimental setting. \algname-m refers to the medium variant.}
\vspace{-8mm}
\label{tab:mmd}
\end{table}

\subsubsection{World Simulator}
We have further validated our model's capability for scene extrapolation as a world simulator on the Waymo Sim Agents Benchmark. As shown in Table~\ref{tab:wodsimagent_test}, our method significantly outperforms competitors in interactive, map-based metrics, and achieves the lowest overall minADE, marking it as state-of-the-art in terms of realism meta metric. The result strongly affirms the exceptional realism and interactivity of our method as a simulator. Moreover, it paves the way for providing a more realistic and interactive environment for various downstream applications.

\begin{table*}[h]
\begin{center}
    \resizebox{0.7\textwidth}{!}{%
        \begin{tabular}{l|cc|ccc}
        \hline
        \multirow{2}{*}{Agent Policy} & \multicolumn{1}{c}{\cellcolor{Gray}Meta Metrics$\uparrow$} & \multirow{2}{*}{minADE$\downarrow$}  & \multicolumn{1}{c}{Kinematic Metrics$\uparrow$} & \multicolumn{1}{c}{Interactive Metrics$\uparrow$} &
        \multicolumn{1}{c}{Map-based Metrics$\uparrow$}
        \\
       & \multicolumn{1}{c}{\cellcolor{Gray}} &  &  &  &  \\
       \hline
Logged Oracle & \cellcolor{Gray}0.7220  & 0.000     & 0.4857 & 0.8415& 0.9043    \\
Constant Velocity & \cellcolor{Gray}0.2870 & 7.923 & 0.0465 & 0.4087 & 0.4547   \\
SBTA-ADIA \cite{Mo23tr_SBTA_ADIA} &  \cellcolor{Gray}0.4202 & 3.611  & 0.3574 & 0.4283 & 0.5087  \\
\textsc{CAD} \cite{Chiu23tr_CollisionAvoidanceDetour} &  \cellcolor{Gray}0.5314 & 2.315  & 0.3357 & 0.5638 & 0.7688   \\
Joint-Multipath++\cite{Wang23tr_JointMultipathSimAgents} & \cellcolor{Gray}0.5330  & 2.049  & 0.4078 & 0.5728 & 0.6677   \\
Wayformer\cite{Nayakanti23icra_Wayformer} & \cellcolor{Gray}0.5750 & 2.498  & 0.3120 & 0.7048 & 0.7747     \\
MTR+++\cite{Qian23tr_SimpleEffectiveSimMultiAgent} &  \cellcolor{Gray}0.6077 & 1.682 & 0.3597 & 0.7172 & 0.8151  \\
MVTA \cite{wang2023multiverse} & \cellcolor{Gray}0.6361 & 1.869  & 0.4175 & 0.7390 & 0.8139  \\
Trajeglish~\cite{philion2023trajeglish} & \textbf{\cellcolor{Gray}0.6437} & 1.615 & 0.4157 & 0.7646 & 0.8104 \\
MVTE$^{\star}$\cite{wang2023multiverse}   &  \textbf{\cellcolor{Gray}0.6448} & 1.677  & \textbf{0.4202} & 0.7506 & 0.8271   \\

\textbf{\algname-m} & \textbf{\cellcolor{Gray}0.6432} & \textbf{1.590} & 0.3994 & \textbf{0.7657} & \textbf{0.8290}\\

        \hline
        \end{tabular}
    }
\end{center}
\caption{\emph{Test} set result of WOD Sim Agents Benchmark. Note: $\star$ indicates the use of model ensemble techniques. Methods are ranked by composite metric on the \href{https://waymo.com/open/challenges/2023/sim-agents/}{V1 Leaderboard}. Our method significantly outperforms others in interactive, map-based metrics, and achieves the lowest overall minADE, marking it as state-of-the-art in terms of realism meta metric. The highest score is highlighted in bold. Detailed results for each component metric and their descriptions are provided in the Appendix.}
\label{tab:wodsimagent_test}
\vspace{-8mm}
\end{table*}

\subsubsection{Interactive Planning}
To validate the effectiveness of \algname as a world simulator in interactive planning task, we conducted both open-loop and closed-loop experiments on the large-scale planning dataset, nuPlan. In the open-loop experiments, we treated planning as an ego-prediction task. By iteratively unrolling the predicted ego states, we can obtain possible future trajectories of the ego vehicle in an interactive environment. To eliminate randomness, we parallel sampled $N$ times and averaged these $N$ rollouts trajectories. As shown in Table~\ref{tab:open-loop}, compared to other methods, our method achieved the best results on most open-loop metrics, demonstrating the feasibility of imitating a planning policy by learning a world simulator.

\begin{table*}[h]

\label{table:nuplan}
\begin{center}
\resizebox{0.7\textwidth}{!}{
\begin{tabular}{l|cccccc}
\hline
\multicolumn{1}{l|}{\bf Methods} & \cellcolor{Gray}{Score $\uparrow$} & 8sADE $\downarrow$ & 3sFDE $\downarrow$ & 5sFDE $\downarrow$ & 8sFDE $\downarrow$ & MR $\downarrow$ \\ 
\hline
IDM~\cite{treiber2000congested}& \cellcolor{Gray}37.7 & 9.600 & 6.256 & 10.076 & 16.993 & 0.552  \\
PlanCNN~\cite{renz2022plant}& \cellcolor{Gray}64.0  & 2.468 & 0.955 & 2.486 & 5.936 & 0.064 \\
Urban Driver~\cite{scheel2022urban}& \cellcolor{Gray}76.0 & 2.667 & 1.497 & 2.815 & 5.453 & 0.064  \\
PDM-Open~\cite{Dauner2023CORL}& \cellcolor{Gray}85.8 &2.375 & 0.715 & 2.06 & 5.296 & 0.042  \\
CKS-124m~\cite{sun2023large}&\cellcolor{Gray}88.0 & \textbf{1.777} & 0.951 & 2.105 & 4.515 & 0.053  \\
CKS-1.5b~\cite{sun2023large}& \cellcolor{Gray}86.6 & 1.783 & 0.971 & 2.140 & 4.460 & 0.047 \\
\textbf{\algname-m}& \cellcolor{Gray}\textbf{88.6}  & 1.820 & \textbf{0.743} & \textbf{1.833} & \textbf{4.453} & \textbf{0.046} \\
\hline
\end{tabular}
}
\end{center}
\caption{Performance comparison of open-loop planning on the validation 14 set of the nuPlan dataset. Our model outperforms all the other models across the majority of metrics.}
\label{tab:open-loop}
\vspace{-8mm}
\end{table*}

We further integrated the imitation policy with a set of rule-based policies, enhancing it with a interactive scoring module powered by \algname, and validated its effectiveness in a closed-loop reactive environment. We have compared it with previous state-of-the-art results on the test14-hard splits~\cite{renz2022plant}, as shown in Table~\ref{tbl:close-loop}. Additionally, we have benchmarked our planner and other competitors in two different reactive environments: IDM and \algname. The experimental results demonstrate that our policy outperformed all the others in both reactive environments, showing improvements in progress, drivable area compliance, and time to collision, among other metrics, when compared to the previous state-of-the-art, PDM~\cite{Dauner2023CORL}. The reliance of PDM on a constant velocity assumption results in overly conservative decision-making. In contrast, our adoption of a stochastic modeling approach allows for better handling of future uncertainties, thereby achieving a better balance between progress and safety (TTC).

\begin{table*}[h]
\begin{center}
    \resizebox{0.9\textwidth}{!}{%
        \begin{tabular}{c|c|c|c|c|c|c|c|c|c|c}
        \hline
        \multirow{2}{*}{Policy} & \multicolumn{1}{c|}{Reactive} & \multicolumn{1}{c|}{\cellcolor{Gray}Driving} & \multirow{2}{*}{Col.} & \multicolumn{1}{c|}{Driv.} & \multicolumn{1}{c|}{Dir.} & \multicolumn{1}{c|}{Making} & \multirow{2}{*}{TTC} & \multicolumn{1}{c|}{Speed} & \multirow{2}{*}{Progress} & \multirow{2}{*}{Comfort} \\
               & \multicolumn{1}{c|}{env}      & \multicolumn{1}{c|}{\cellcolor{Gray}Score} &  & \multicolumn{1}{c|}{Comp.} & \multicolumn{1}{c|}{Comp.} & \multicolumn{1}{c|}{Progress} &  & \multicolumn{1}{c|}{Limit} &  &  \\
       \hline
       Expert~\cite{jcheng2023plantf} & IDM & \cellcolor{Gray}68.80&-& -& -& -& -& -& -& -\\
       UrbanDriver~\cite{scheel2022urban}& IDM &  \cellcolor{Gray}49.07&-& -& -& -& -& -& -& -\\
       GC-PGP~\cite{hallgarten2023prediction}& IDM & \cellcolor{Gray}39.63&-& -& -& -& -& -& -& -\\
       GameFormer~\cite{Huang_2023_ICCV}& IDM &\cellcolor{Gray}68.83&-& -& -& -& -& -& -& - \\
       planTF~\cite{jcheng2023plantf} & IDM & \cellcolor{Gray}61.70&-& -& -& -& -& -& -& -\\
       hoplan~\cite{hu2023imitation} & IDM & \cellcolor{Gray}75.06 & 89.33 & 94.85 & 96.13 & \textbf{97.05} & 80.51 & 95.28 & 85.02 & \textbf{98.52} \\	
       PDM~\cite{Dauner2023CORL} & IDM & \cellcolor{Gray}75.18& \textbf{95.22}&	95.58&	99.08&	93.38&	84.19&	\textbf{99.53}&	75.47&	83.45  \\ 
       \textbf{\algname hybrid} & IDM& \textbf{\cellcolor{Gray}77.77}&	94.36&	\textbf{98.98}&	98.95&	94.41&	\textbf{87.46}&	97.51 & \textbf{77.08} &	79.84  \\
       \hline
        hoplan~\cite{hu2023imitation} & \algname & \cellcolor{Gray}69.56 & 93.63 & 96.25 & 97.37& 86.14 & 88.01 & 94.17 & \textbf{76.74} & \textbf{98.87} \\ 
        PDM~\cite{Dauner2023CORL} & \algname&  \cellcolor{Gray}72.26 &	\textbf{97.00} &	95.50&	99.25&	88.38&	86.89&	\textbf{99.51}&	70.43&	85.39 \\ 
        \textbf{\algname hybrid} &  \algname& \textbf{\cellcolor{Gray}73.60} &	94.98&	\textbf{98.97}&	99.31&	\textbf{88.76}&	\textbf{89.82}&	97.18&	72.76&	82.00  \\
        \hline
        \end{tabular}
        }
\end{center}
\caption{Results on the Hard-14 Test Set: Higher Scores Indicate Better Performance Across All Metrics. Our driving score in both IDM and \algname evaluation environments are superior to other competitors, and we achieve a better balance between progress and safety (TTC). For detailed metrics definition, please refer to Appendix~\ref{sec:nuplan_metrics}. In this context, "Col." refers to collision metric, "Driv. Comp." refers to drivable area compliance metric, and "Dir. Comp." refers to driving direction compliance, and "TTC" stands for time to collision.}
\label{tbl:close-loop}
\vspace{-8mm}
\end{table*}


\subsubsection{Online Training}
We further demonstrate the effectiveness of using \algname as an online training environment, as detailed in Table~\ref{tbl:ablation RL}. For comparison, we employed Imitation Learning (IL) and the Soft Actor-Critic (SAC) algorithm for online training, with all experiments utilizing a consistent ResNet18 as the policy model. 
We selected both logsim~\cite{chen2021learning} and \algname as our environments for online training and testing, respectively. Comparing exp ID 1, 2 with ID 3, 4, we can conclude that using \algname as the training environment significantly enhances the RL model's performance across different testing environments, substantially reducing the collision rate and improving progress, and outperforms the IL baseline. Moreover, policies trained in the \algname environment achieved better results in realistic, reactive \algname testing environments compared to those trained in logsim environments. These improvements are attributed to \algname's capability to generate a vast amount of realistic, interactive, and diverse training scenarios, greatly enhancing the efficiency of RL training and reducing the sim-to-real domain gap. These findings highlight the effectiveness of \algname as an environment for online training.

\begin{table*}[h]
\begin{center}
    \resizebox{0.8\textwidth}{!}{%
        \begin{tabular}{c|c|c|c|c|c|c|c|c}
        \hline
        \multirow{2}{*}{ID} & \multirow{2}{*}{Policy} & \multicolumn{1}{c|}{Training} & \multicolumn{1}{c|}{Testing} & \multicolumn{1}{c|}{\cellcolor{Gray}Driving} & \multicolumn{1}{c|}{Col.} & \multicolumn{1}{c|}{Out of Driv.}  & \multicolumn{1}{c|}{Progress} & \multicolumn{1}{c}{Comfort} \\

             & & \multicolumn{1}{c|}{env}  & \multicolumn{1}{c|}{env}      & \multicolumn{1}{c|}{\cellcolor{Gray}Score $\uparrow$} & \multicolumn{1}{c|}{Rate$\downarrow$} & \multicolumn{1}{c|}{Rate$\downarrow$} & \multicolumn{1}{c|}{meters$\uparrow$} &  \multicolumn{1}{c}{Score$\uparrow$} \\
       \hline
       0 & IL & log data &  logsim&  \cellcolor{Gray}0.595	& 0.248 &	\textbf{0.014}&	44.69&	0.892 \\
       1& SAC & logsim & logsim & \cellcolor{Gray}0.587	& 0.259	&0.062&	49.77	& \textbf{0.923}\\
       2& SAC & \algname & logsim & \textbf{\cellcolor{Gray}0.643}&	\textbf{0.193}	&0.077	&\textbf{56.78}	&0.845 \\
        \hline
       3& SAC & logsim & \algname & \cellcolor{Gray}0.569 &	0.246	&0.061 	& 44.46 & \textbf{0.920}	\\ 
       4& SAC & \algname & \algname & \textbf{\cellcolor{Gray}0.626}&	\textbf{0.215}&	0.061 &	\textbf{53.58}&	0.862   \\
        \hline
        \end{tabular}
        }
\end{center}
\caption{
In an ablation study on the nuPlan random 1000 splits validation set, we observed significant improvements when using \algname for online RL training across different environments, compared to non-reactive logsim environments. This enhancement is evident in both logsim and \algname testing environments. Please note that the metric definition utilized here  is more strict than the original nuPlan metrics. For detailed metric definitions, see the Appendix~\ref{sup: rl-metric}. } 
\vspace{-8mm}
\label{tbl:ablation RL}
\end{table*}

\subsubsection{Scaling Laws}
\label{sec: scaling laws}
\begin{figure}[ht]
    \vspace{-5mm}
  \centering
  \begin{minipage}{0.4\textwidth}
    \includegraphics[width=\linewidth]{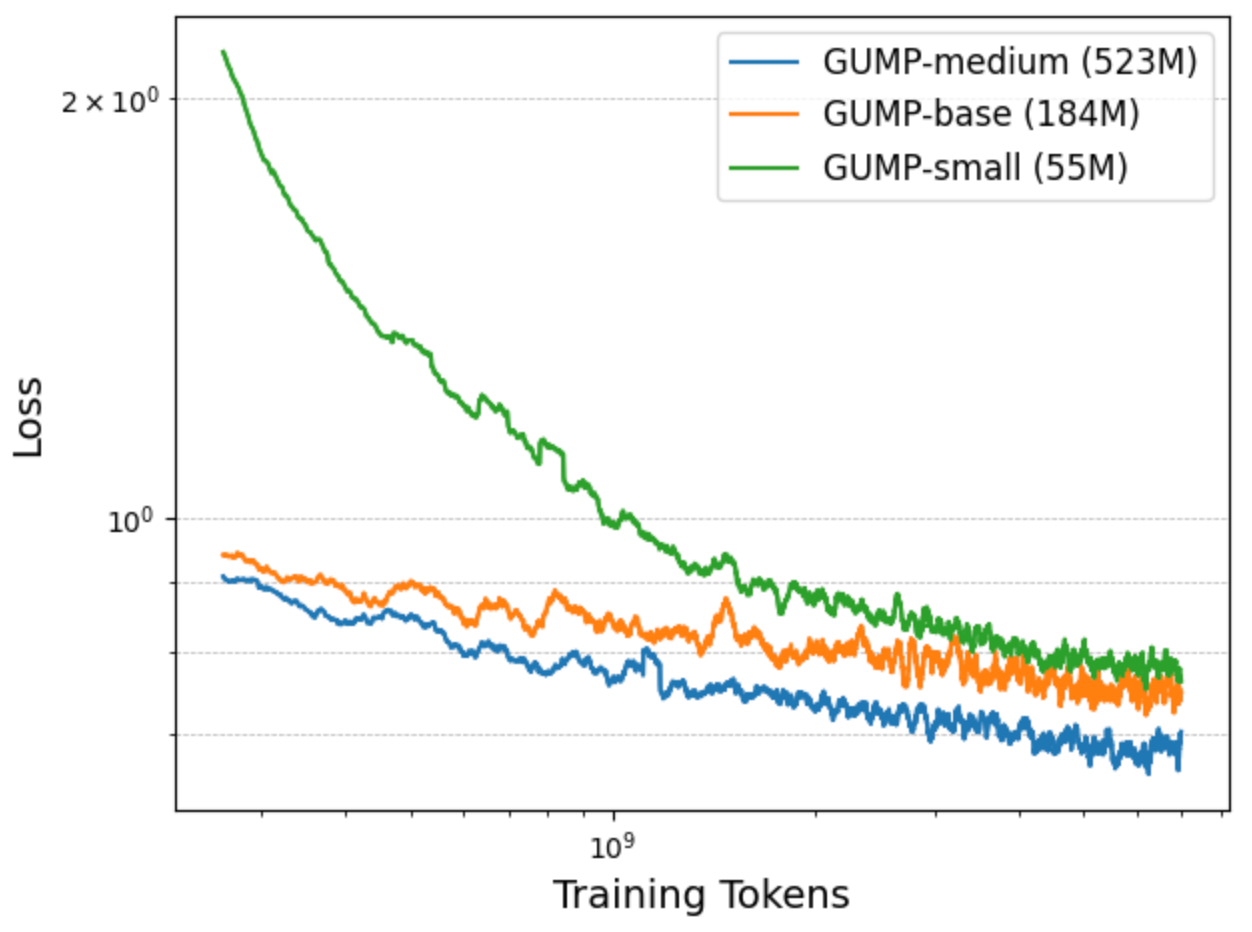} 
  \end{minipage}\hfill
  \begin{minipage}{0.44\textwidth}
    \includegraphics[width=\linewidth]{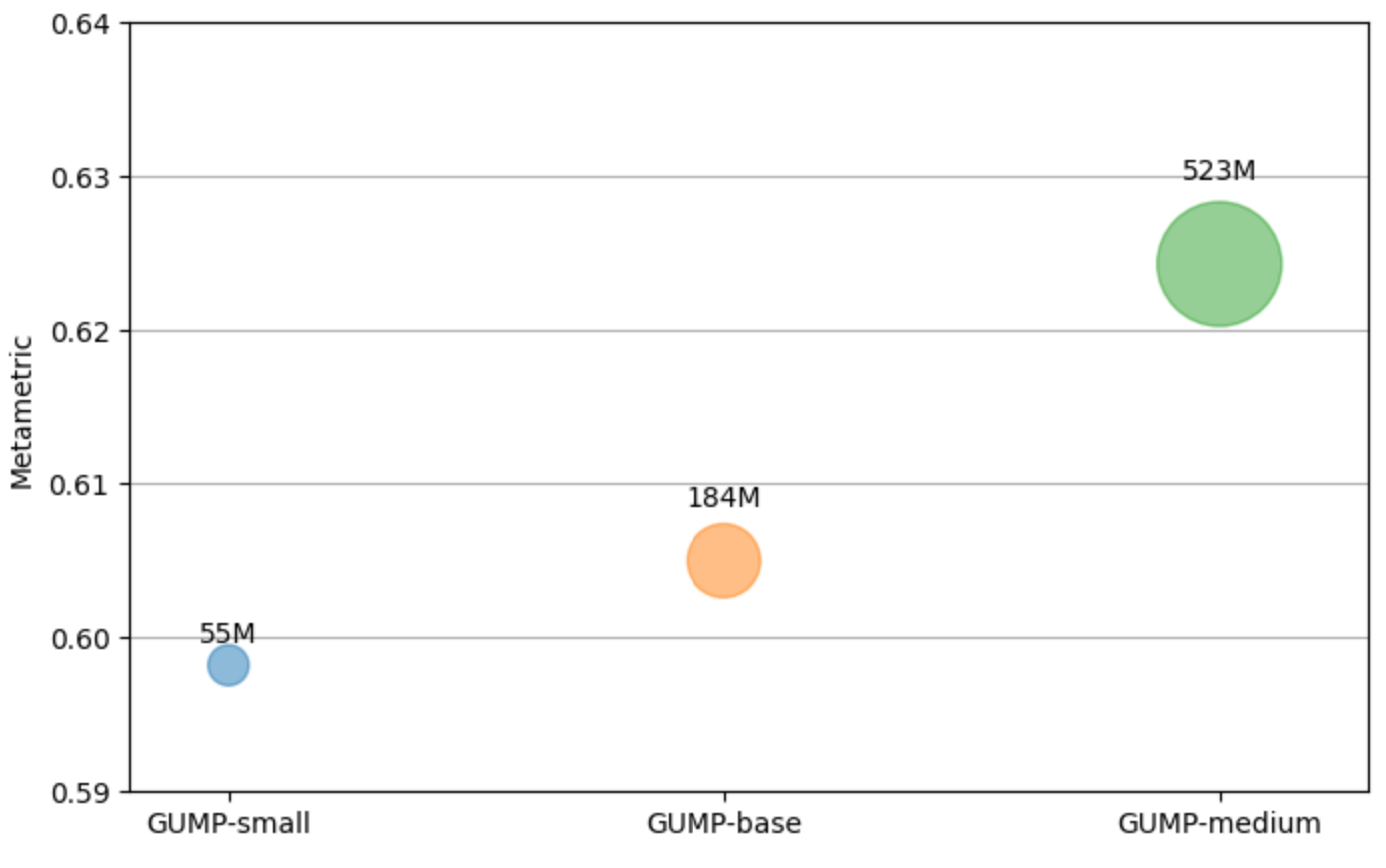} 
  \end{minipage}
  \caption{Left: Training loss versus number of training tokens consumed, with axes shown on a logarithmic scale. Right: WOD Sim Agents meta metric, where a larger area indicates larger model parameters.}
  \label{fig:scaling}
\vspace{-12mm}
\end{figure}

As we adopted the same paradigm as large language models (LLMs), which are transformer-based, and both aim at predicting the next token as the training objective, we have the reason to believe that our model also possesses strong scalability. To validate this, we trained three variants of our model with different MCT backbone capacities, i.e \algname-small, \algname-base and \algname-medium, and measured the cross-entropy loss versus the number of tokens consumed during training, as well as the simulation realism meta metric in WOD Sim Agents validation set, as illustrated in Fig.~\ref{fig:scaling}. We can observe that our model demonstrates high scalability, as both training FLOPs and model capacity increasing. Furthermore, an increase in model capacity leads to a significant reduction in training loss and a notable enhancement in validation meta metric, especially when contrasting the \algname-medium with the \algname-base model. This phenomenon suggests that our model still harbors considerable potential for enhancement, as indicated by the scaling law~\cite{kaplan2020scaling, hu2023gaia}.

\section{Conclusion and Future Work}
In this work, we have proposed \algname, a unified generative model designed to solve a broad spectrum of motion planning tasks in the domain of autonomous driving. Our model has demonstrated exceptional scalability and effectiveness across various downstream applications, showcasing its potential to serve as a foundation model in this field.

\textbf{Limitations and Future Directions:} Our model, while effective, has potential for further refinement. Firstly, there is room to enhance the model's efficiency through engineering efforts such as model quantization and key-value caching. Additionally, adopting vectorized map inputs instead of raster map inputs may potentially yield more accurate map information. Furthermore, our approach primarily focuses on the structured dynamics of driving scenarios. Expanding our model to integrate sensor data directly would be a logical next step, enabling truly end-to-end operation and potentially enhancing its applicability and performance in real-world settings.

\bibliographystyle{splncs04}
\bibliography{egbib}


\newpage
\appendix
\clearpage
\newcommand\appendixsection[2]{
    \noindent\textbf{#1} \dotfill #2\par
}

\newcommand\appendixsubsection[2]{
    \noindent\hspace*{1em}#1 \dotfill #2\par
}

\section*{Appendix}
\hypersetup{linkcolor=black}
\appendixsection{A \hyperref[ape:A]{Related Works}}{21}
\appendixsubsection{A.1 \hyperref[ape:A1]{Scene Generation}}{21}
\appendixsubsection{A.2 \hyperref[ape:A2]{Reactive Simulation}}{21}
\appendixsubsection{A.3 \hyperref[ape:A3]{Interactive Planning}}{22}
\appendixsubsection{A.4 \hyperref[ape:A4]{Online Learning with RL}}{22}
\appendixsection{B \hyperref[sec: dataset and benchmark]{Dataset and Metrics}}{23}
\appendixsubsection{B.1 \hyperref[ape:B1]{Dataset}}{23}
\appendixsubsection{B.2 \hyperref[ape:B2]{Metrics}}{23}
\appendixsection{C \hyperref[sec: experimrnt setup]{Experimental setup}}{28}
\appendixsubsection{C.1 \hyperref[ape:C1]{Scene Generation}}{28}
\appendixsubsection{C.2 \hyperref[ape:C2]{World Simulator}}{28}
\appendixsubsection{C.3 \hyperref[ape:C3]{Interactive Planning}}{29}
\appendixsubsection{C.4 \hyperref[ape:C4]{Online Training}}{29}
\appendixsection{D \hyperref[ape:D]{Training setup}}{30}
\appendixsubsection{D.1 \hyperref[ape:D1]{Raster Input}}{30}
\appendixsubsection{D.2 \hyperref[ape:D2]{Data Augmentation}}{30}
\appendixsubsection{D.3 \hyperref[ape:D3]{Loss}}{31}
\appendixsubsection{D.4 \hyperref[ape:D4]{Hyperparameters}}{31}
\appendixsection{E \hyperref[sec: ablation study]{Ablation Study}}{32}
\appendixsubsection{E.1 \hyperref[sup: nar ablation]{Effectiveness of NAR conversion}}{32}
\appendixsubsection{E.2 \hyperref[ape:E2]{Effectiveness of Prediction Chunking and Temperal Aggregation}}{32}
\appendixsubsection{E.3 \hyperref[ape:E3]{Ablation Study of Decay Rate $\gamma$}}{33}
\appendixsubsection{E.4 \hyperref[ape:E4]{Ablation Study of Temperature}}{34}
\appendixsubsection{E.5 \hyperref[ape:E5]{Ablation Study of the Number of Conditioned Frames}}{34}

\appendixsection{F \hyperref[sup: quanlitative]{Qualitative Analysis}}{35}
\appendixsubsection{F.1 \hyperref[ape:F1]{Scene Generation}}{36}
\appendixsubsection{F.2 \hyperref[ape:F2]{Diverse future}}{36}
\appendixsubsection{F.3 \hyperref[sup: reactive simulation]{Reactive Simulation}}{37}

\appendixsection{G \hyperref[sup: trajectory]{WOD Motion Results}}{38}
\appendixsubsection{G.1 \hyperref[ape:G1]{WODM validation set}}{38}
\appendixsubsection{G.2 \hyperref[ape:G2]{Per-type Results of WOMD Validation}}{39}

\appendixsection{H \hyperref[ape:H]{Per-component WOD Sim Agent Metric}}{39}

\newpage



\hypersetup{linkcolor=red}
\section{Related Works}
\label{ape:A}

\subsection{Scene Generation}
\label{ape:A1}

As a supplement to real data, generated scenarios establish the initial conditions for reactive simulations. Previous efforts have involved the creation of agents guided by heuristic rules~\cite{yang1996microscopic, dosovitskiy2017carla, lopez2018microscopic, prakash2019structured} or fixed grammars~\cite{kar2019meta, devaranjan2020meta}. However, the manual generation of such scenarios requires extensive human labor and struggles to achieve the necessary realism, diversity, and generalization for downstream tasks, especially in new environments. Recent work has started to adopt an end-to-end fully learned approach, either through an autoregressive architecture that places agents one by one~\cite{tan2021scenegen, feng2023trafficgen} or a diffusion-based architecture~\cite{pronovost2024scenario}. Yet, no work has exploited a more scalable or unified transformer-based language model for controllable scene generation. Moreover, our method allows for a coarse control over the entire scene through a global description prompt and a fine control by directly modifying each agent's states, e.g. enforcing traffic rule constraints. These features enable our model to strike a better balance between diversity and controllability throughout the generation process.

\subsection{Reactive Simulation}
\label{ape:A2}

Traditional simulators are built on computer graphics and human prior knowledge, including physical laws, lighting conditions, and hand-crafted traffic dynamics~\cite{dosovitskiy2017carla, lopez2018microscopic, li2022metadrive, caesar2022nuplan}. However, relying on a vast collection of manually created scene assets and heuristic driving policies for intelligent agents not only introduces a significant domain gap but also results in a lack of diverse scenarios and realism.

Recently, there has been a growing emphasis on building world simulators as generative models through a data-driven approach~\cite{santana2016learning,yan2021videogpt, hafner2020mastering, hu2023gaia, ha2018worldmodels, li2023drivingdiffusion, zhang2023learning}. Compared to those end-to-end models, structured input provides a simpler and more efficient setting, facilitating a variety of downstream applications and onboard deployments. Predictive models, based on the unfolding of agents' states over time, fall into open-loop and closed-loop categories. In an open-loop setting, each agent makes decisions—either marginally~\cite{varadarajan2022multipath++, shi2022motion, gu2021densetnt, hu2023_uniad} or jointly~\cite{ngiam2021scene, shi2024mtr++, hu2022hope}—based solely on historical information and generates predictions for their trajectory over a short future period. In contrast, closed loop simulations incorporate feedback mechanisms, where the decisions of each agent are influenced by the current state of the system, including the actions of other agents. This leads to a more dynamic, interactive and realistic simulation environment, where agents continuously update their decisions based on the evolving scenario, such as~\cite{suo2021trafficsim, seff2023motionlm}. Notably, WOSAC~\cite{montali2023waymo} established the first evaluation benchmark for closed-loop simulators, attracting numerous participants~\cite{wang2023multiverse, philion2023trajeglish, Qian23tr_SimpleEffectiveSimMultiAgent, Nayakanti23icra_Wayformer}. As a closed-loop simulator, our work is most related to MotionLM~\cite{seff2023motionlm} and Trajenglish~\cite{philion2023trajeglish}, as both employ a GPT-like autoregressive predictive model. However, in contrast to them, which tokenize the action space, we employ a ``key-value pair'' tokenization strategy and directly quantize the state space. Through this method, we can achieve Non-Autoregressive (NAR) transitions within frames, significantly speeding up inference, and endowing the model with generative capabilities. Moreover, it has the flexibility to handle the disappearance and emergence of agents. These capabilities foster a wider range of downstream applications, which require higher scalability, efficiency, and flexibility.

\subsection{Interactive Planning}
\label{ape:A3}
The ability to effectively model the interactions between road users and autonomous vehicles is crucial for enhancing the safety and comfort of self-driving technology. 
An optimal policy planning algorithm should enable multi-stage reasoning, incorporating bidirectional interactions between the agent and it's environment.  Previous research has tackled this challenge through two main approaches: Some studies have employed neural networks to implicitly and iteratively capture the interactions between the ego vehicle and other road users~\cite{Huang_2023_ICCV,shi2022motion,shao2023reasonnet,hu2023_uniad,chen2023deepemplanner}. Others have taken a more explicit route, utilizing model predictive control (MPC) in conjunction with tree search expansion to navigate complex interactions~\cite{tesla2022aiday,chen2022interactive,chen2023tree,huang2023dtpp}.
Among these approaches, certain studies~\cite{chen2022interactive,chen2023tree, huang2023dtpp} have simplified the modeling of bidirectional interaction by combining a simple non-reactive rule-based planner with a model-based, ego-conditioned predictor. In contrast, PDM~\cite{Dauner2023CORL} simplifies interaction by assuming a non-reactive environment with constant velocity agents, and a reactive Intelligent Driver Model (IDM)-based~\cite{treiber2000congested} ego-planner. Our work advances beyond these methodologies by coupling our realistic ego-conditioned simulator with a reactive planner, adaptable to any rule-based or neural network-driven ego-planner. This approach allows for a more accurate modeling of interactions and the stochastic nature of future scenarios.

\subsection{Online Learning with RL}
\label{ape:A4}
The reinforcement learning (RL) domain has witnessed considerable progress, fostering algorithms adept at solving various types of tasks via mode-free~\cite{haarnoja2018soft, schulman2017proximal} and model-based approaches~\cite{hafner2020mastering,hafner2023mastering}. 
The progress in RL has contributed to the paradigm shift in autonomous driving
from open-loop trajectory prediction-based approach to closed-loop interaction-based approach.
One critical component to enable closed-loop training is a reactive environment, generating the state of the next time step given the current state and action taken~\cite{chen2021learning, mile2022, pan2022iso, gao2022enhance}. Nevertheless, the simplicity of these models often results in environments that fall short of realism and interactivity, leading to a pronounced simulation-to-reality gap in complex scenarios. Addressing this, some research focuses on the creation of more realistic and intelligent training environments~\cite{feng2023trafficgen,li2022metadrive}. Different from previous approaches, we introduce a scalable, data-driven model capable of imitating interactive agent behaviors and generating realistic, diverse driving scenarios. Its efficiency aligns well with the demands of online RL training. Such an environment could bridge the simulation-reality gap, and push RL closer to practical application in autonomous driving.

\section{Dataset and Metrics}
\label{sec: dataset and benchmark}
\subsection{Dataset}
\label{ape:B1}
\subsubsection{The WOD Motion Dataset}
We have utilized the Waymo Open Motion Dataset (WOMD) to train and evaluate for scene generation, reactive simulation and motion prediction tasks. WOMD contains 7.64 million unique tracks from 574 driving hours across 1750 km urban roadways collected in six cities of the United States. Specifically, the WOMD v1.2.0 release includes 486,995 train, 44,097 validation, and 44,920 test scenarios. Each scenario consists a 9.1 seconds 10 Hz driving log, partitioned into 11 history frames and 80 future frames. All reported variants of our model is trained on the full training dataset. To further augment the data volume, we treat different objects as the center object, ultimately yielding approximately 2.6 million training scenarios. 

\subsubsection{The nuPlan Dataset}
We utilized the nuPlan Dataset \cite{caesar2022nuplan} for scenario prompt conditioned scene generation, interactive planning and online training experiments. The dataset encompasses 1,500 hours of diverse driving scenarios from urban environments such as Las Vegas, Boston, Pittsburgh, and Singapore. These scenarios are annotated with various traffic situations including merges, lane changes, interactions with cyclists and pedestrians, among others. For training, a random sampling strategy is adopted, with 50,000 scenarios selected for each scenario type from the training subset, culminating in approximately 1.5 million scenarios. 

\subsection{Metrics}
\label{ape:B1}
\subsubsection{Scene Generation Metrics}

To measure the quality of generated scenarios, we treat the generated agents as samples from a distribution, and compare it with ground-truth agents, which are also treated as samples from another distribution. Similar to Trafficgen~\cite{feng2023trafficgen}, we use maximum mean discrepancy (MMD) as the metric to measure the difference between the generated distribution and the corresponding ground-truth distribution. Given two distributions \( p \) and \( q \) with Gaussian kernel \( k \), the maximum mean discrepancy is defined as

\begin{equation}
\begin{aligned}
MMD^2(p, q) = \mathbb{E}_{x,x' \sim p}[k(x, x')] + \mathbb{E}_{y,y' \sim q}[k(y, y')] - 2\mathbb{E}_{x \sim p, y \sim q}[k(x, y)]
\end{aligned}
\end{equation}

We calculate this metric for generated attributes including agent center positions in \( \mathbb{R}^2 \), agent heading angle in \( \mathbb{R} \), agent velocities in \( \mathbb{R}^2 \), and agent dimensions in \( \mathbb{R}^2 \).

For each pair of generated and ground-truth scenarios, we calcuate MMD for each attributes independently and compute the average value of them. And the final MMD score is averaged across all selected testing dataset.

\subsubsection{The WOD Sim Agents Metrics}
We evaluate our model's simulation capability on the WOMD Sim Agents Benchmark~\cite{montali2023waymo}. Each testing scenario consists a 9.1 seconds 10 Hz driving log, partitioned into 11 history frames and 80 future frames. Trajectories of up to 128 agents (including ego vehicle) are tracked in each scenario. The task is to simulate 32 parallel future rollouts for each agent and scenario in an autoregressive manner. Given an agent in a scenario, the predicted distribution of its future behavior is formed over the 32 rollouts. In order to parameterize the predicted distribution, a total of 3 categories and 9 component metrics are computed:

\begin{itemize}
\item \textbf{Kinematic-based:}
    \begin{itemize}
        \item \textbf{Linear Speed} unsigned magnitude of agent's speed in x, y, z axis.
    
        \item \textbf{Linear Acceleration Magnitude} derivative of linear speed with respect to time in x, y, z axis.
        
        \item \textbf{Angular Speed} signed minimum difference between consecutive angular heading
        
        \item \textbf{Angular Acceleration Magnitude} derivative of angular speed with respect to time
    \end{itemize}
    
\item \textbf{Interaction-based:}
    \begin{itemize}
        \item \textbf{Distance to nearest object} for each agent represented by a box polygon, the signed distance from the nearest object in the scenario.
        \item \textbf{Collisions} Count of collisions with other objects, recognized when \textbf{Distance to nearest object} is negative.
        \item \textbf{Time-to-collision} Time takes before an agent collide with the agent it is following, assuming constant speed.
    \end{itemize}

\item \textbf{Map-based:}
    \begin{itemize}
        \item \textbf{Distance to road edge} Signed distance to the nearest road edge, where a road egde is represented as a 2D vector.
    
        \item \textbf{Road departures} Indicator of an agent going off road at any time.
    \end{itemize}
\end{itemize}

 A component score is calculated by testing the Negative Log Likelihood of the ground truth agent behavior over the corresponding distribution, masked by validity  $v_t$: $m = \exp\Big(- \frac{1}{\sum\limits_t \mathbbm{1}\{v_t\} } \sum\limits_{t} \mathbbm{1}\{v_t\} NLL_t\Big)$. Each component metric is assigned a weight, and each categorical metric is computed by taking the weighted average of all corresponding component metrics: $\mathcal{M}_{c} = \frac{\sum_{i} w_i m_i}{\sum_{j} w_j}$. We report all categorical metrics on the test dataset in Tab.~\ref{tab:per-metric-scores-test}. The final composite metric is calculated as a weighted average over all component metrics. Our reported model performance is based on the \href{https://waymo.com/open/challenges/2023/sim-agents/}{V1 Leaderboard}, which improved collision and offroad calculation from the previous \href{https://waymo.com/open/challenges/2023/sim-agents-v0/}{V0 Leaderboard}. 

\subsubsection{The nuPlan Planning Metrics}
\label{sec:nuplan_metrics}

Our evaluation concentrates on the model's planning abilities in closed-loop with reactive agents (CL-R) task. In these setups, the planner generates a trajectory at each timestep, which is used as a reference by the controller to incrementally adjust the vehicle's state. In CL-R tasks, we apply two distinct world models for the surrounding agents: a rule-based IDM environment and a model-based environment powered by \algname, both controls all non-ego agents dynamically.

The planning score, as defined by the equation \eqref{equ: reward}, is reported on on the test dataset. We exactly follow the nuPlan benchmark metrics~\cite{caesar2022nuplan}, which reflects the model's performance across safety, efficiency, and comfort metrics in CL-R tasks. This score is aggregated by selected metrics for the driven trajectory generated from the planner. The aggregation function is a hybrid hierarchical-weighted average function of individual metric scores. 

The first part $\Theta$ of equation \eqref{equ: reward} gets a zero score for a driven scenario if any one of those following situations happens. 
\begin{itemize}
\item There is an at\_fault collision with a vehicle or a VRU (pedestrian or bicyclist);
\item There are multiple at\_fault collisions with objects (e.g. a cone);
\item There is a drivable\_area violation;
\item Ego drives into uncoming traffic more than 6 m;
\item Ego is not making enough progress.
\end{itemize}

The second part $\Phi$ of equation \eqref{equ: reward} is a weighted average of other selected metrics’ scores. Those selected metrics and their corresponding weight are listed as following.

\begin{table}[ht]
\centering

\begin{tabular}{>{\raggedright\arraybackslash}p{5.5cm}|>{\centering\arraybackslash}p{2.cm}} 
\toprule
Metric Name & $\omega_{n}$  \\
\midrule
driving\_direction\_compliance &  5 \\
time\_to\_collision\_within\_bound & 5 \\
speed\_limit\_compliance & 4 \\
ego\_progress\_along\_expert\_route & 5 \\ 
Ego\_is\_comfortable & 2 \\

\bottomrule
\end{tabular}
\caption{Selected metrics for the second part $\Phi$ of equation \eqref{equ: reward} .}
\label{tab:WeightedMetrics}
\end{table}

The following component metrics contribute to the overall planning score:
\begin{itemize}
\item \textbf{No at-fault Collisions} A collision is defined as the event of ego’s bounding box intersecting another agent’s bounding box. To define the collision score for a scenario, we only consider collisions that should have been prevented if planner performed properly. For simplicity, we call these collisions at-fault. 

\item \textbf{Drivable area compliance} Ego should drive in the mapped drivable area at all times. Drivable area compliance metric identifies the frames when ego drives outside the drivable area. Due to over-approximation of ego’s bounding box, we allow for a small infringemenet outside the drivable area (max\_violation\_threshold = 0.3m).

\item \textbf{Driving direction compliance} This metric is defined to penalize ego when “it drives into oncoming traffic”. The metric computes the movement of ego’s center during a 1 second time\_horizon along the driving direction defined according to the baselines of ego’s lanes or lane-connectors. The score is set to 1 if it does not drive/move against the flow more than driving\_direction\_compliance\_threshold (= 2 m) and 0 if it drives against the flow more than driving\_direction\_violation\_threshold (= 6 m), and 0.5 otherwise.

\item \textbf{Ego progress along the expert’s route ratio} This metric is used to evaluate progress of the driven ego trajectory in a scenario by comparing its progress along the route that expert takes in that scenario.

\item \textbf{Making progress} This metric is defined as a boolean metric based on the “Ego progress along the expert’s route ratio”. It’s score is set to 1 if the ratio is more than the selected threshold (min\_progress\_threshold = 0.2), and is set to 0 otherwise.

\item \textbf{Time to Collision (TTC) within bound} TTC is defined as the time required for ego and another track to collide if they continue at their present speed and heading.

\item \textbf{Speed limit compliance} This metric evaluates if ego’s speed exceeds the associated speed limit in the map. Speed limit violation at each frame is defined based on the difference between ego’s speed and the speed limit, if ego’s speed is higher than the speed limit (over-speeding). 

\item \textbf{Comfort} We measure the comfort of ego’s driven trajectory by evaluating minimum and maximum longitudinal accelerations, maximum absolute value of lateral acceleration, maximum absolute value of yaw rate, maximum absolute value of yaw acceleration, maximum absolute value of longitudinal component of jerk, and maximum magnitude of jerk vector. These variables are compared to thresholds with default values determined empirically from examination of a dataset of expert trajectories.

\end{itemize}

\subsubsection{RL metrics}
\label{sup: rl-metric}

We define safe episodes as those without any critical failures, such as collisions and violations of the drivable area. In the context of RL training, we utilize four specific types of metrics that differ slightly from the nuPlan Planning Metrics mentioned previously in \ref{sec:nuplan_metrics}. These metrics include:
\begin{itemize}
    \item \textbf{Collision Rate} We do not differentiate between at-fault and non-at-fault collisions, setting a higher standard for the planner to avoid all collisions, whether by not colliding with others or by not being collided into. The collision rate is calculated as the percentage of episodes that experience at least one collision.
    \item \textbf{Out of Drivable Area Rate} Diverging from the nuPlan metric, our approach disallows any encroachments outside the drivable area, demanding strict adherence from the planner. The score for drivable area compliance is determined by the percentage of episodes where the drivable area is violated.
    \item \textbf{Progress}  We define a series of equally spaced waypoints along the planned route. Progress is measured by the physical distance the ego vehicle covers towards the next waypoint at each step. The overall progress score represents the average progress made in all safe episodes.
    \item \textbf{Comfort} This metric maintains the same components as defined in the nuPlan Planning Metrics. The comfort score is the proportion of episodes deemed comfortable out of all safe episodes.
\end{itemize}

The final score calculation incorporates Equation \eqref{equ: reward}. For this equation, the $\Theta$ component comprises binary metrics from collision and out-of-drivable-area violations, signifying that these metrics are either 0 (no violation) or 1 (violation occurred). As for the $\Phi$ component, it is determined by first calculating the progress percentage. This is achieved by dividing the actual progress made by the vehicle by an empirical value of 62, which represents the average progress length according to the nuPlan dataset. Subsequently, the average of this progress percentage and the comfort metric is calculated to derive the $\Phi$ value.

\subsubsection{WOD motion metrics}
WOMD's metrics are used to evaluate the motion prediction performance of our model.

\begin{itemize}
\item \textbf{minADE} The minimum Average Displacement Error
computes the L2 norm between groundtruth and the closest joint prediction. 
\item \textbf{minFDE.} The minimum Final Displacement Error is
equivalent to evaluating the minADE at a single time step
T. 
\item \textbf{Overlap rate (OR)} The overlap rate is computed by taking the highest confidence joint prediction from each multimodal joint prediction. If any of the A agents in the jointly
predicted trajectories overlap at any time with any other objects that were visible at the prediction time step (compared
at each time step up to T) or with any of the jointly predicted
trajectories, it is considered a single overlap. The overlap
rate is computed as the total number of overlaps divided by
the total number of predictions. See the supplementary material for details. The overlap is calculated using box intersection, with headings inferred from consecutive waypoint
position differences. 
\item \textbf{Miss rate (MR)} A binary match/miss indicator function
ISMATCH(ˆst, st) is assigned to each sample waypoint at
a time t. The average over the dataset creates the miss
rate at that time step. A single distance threshold to determine ISMATCH is insufficient: we want a stricter criteria
for slower moving and closer-in-time predictions, and also
different criteria for lateral deviation (e.g. wrong lane) versus longitudinal (e.g. wrong speed profile) 
\item \textbf{Mean average precision (mAP)} The Average Precision
computes the area under the precision-recall curve by applying confidence score thresholds ck across a validation
set, and using the definition of Miss Rate above to define
true positives, false positives, etc. Consistent with object
detection mAP metrics, only one true positive is allowed for each object and is assigned to the highest confidence prediction, the others are counted as false positives.
Further inspired by object detection literature, we seek
an overall metric balanced over semantic buckets, some of which may be much more infrequent (e.g., u-turns), so report the mean AP over different driving behaviors. The final mAP metric averages over eight different ground truth
trajectory shapes: straight, straight-left, straight-right, left,
right, left u-turn, right u-turn, and stationary.
\end{itemize}

\section{Experimental setup}
\label{sec: experimrnt setup}

\subsection{Scene Generation}
\label{ape:C1}
For our quantitative experiments in scene generation, we utilized all the validation scenarios from the Waymo Open Motion Dataset (WOMD) version 1.2.0. To ensure a fair comparison with Trafficgen~\cite{feng2023trafficgen},  we followed the testing settings outlined in that research and limited our model to generate vehicles only within a range of -50 meters to 50 meters. Moreover, we omitted scenarios with less than 8 ground-truth vehicles, yielding a total of 28,341 scenarios. We evaluated the first frame of each selected scenario to compute the average score. Under identical testing conditions, we reassessed Trafficgen and found the results to align closely with those reported in the original study. For the scene generation task, we found through experimentation that a slightly higher temperature setting leads to a more diverse distribution of scenes, thereby enhancing the performance of our metrics. Consequently, we selected a temperature setting of 1.25 and a top-k of 200 for this task.

For our qualitative analysis of scene generation, we mainly used the nuPlan Dataset. This dataset was chosen for its diverse and detailed scenario description tags, which are essential for our prompt-based conditioning experiments. These experiments aim to generate scenes that follows the scenario descriptions, offering a more nuanced and targeted approach.

\subsection{World Simulator}
\label{ape:C2}
Our experiments are conducted on the WOD Sim Agents Benchmark, utilizing the WOD Motion Dataset. We adhere to the protocols of the Sim Agents challenge, unrolling 32 futures for each scenario in parallel. Specifically, for the experiments detailed in Tab.\ref{tab:wodsimagent_test}, we utilize the test splits, with evaluation performed by the Waymo Sim Agent Challenge server. For other experiments, such as the Scaling Laws in Sec.~\ref{sec: scaling laws} and various ablations, we employ the sub50 validation dataset, which comprises a total of 1024 scenarios.

Specifically, our model operates in 2Hz, and we interpolate the results to 10Hz for evaluation. For both the Waymo and nuPlan datasets, we use 1 second of history along with the current frame as conditions and predict the information for the next 8 seconds. Our model does not directly output information on the z-axis. To meet the requirements of the Waymo Sim Agents Benchmark, we infer the z value for each agent based on their predicted x and y positions, in conjunction with map data. Specifically, we employ the K-nearest neighbor algorithm to identify the k closest map points in the xy-2D plane location. We then average the z information of these map points to estimate the agent's z-axis position. In our approach, we set k to 4. Additionally, to better combat the jitter caused by the sampling and quantization processes, we employ a simple sliding window algorithm to smooth the final prediction results. This method helps with the stability and the overall smoothness.

\subsection{Interactive Planning}
\label{ape:C3}
Our planning experiments are conducted on the nuPlan Dataset. To develop an interactive planner that builds upon the PDM~\cite{Dauner2023CORL}, we integrate multiple simple policies with an interactive simulator and scorer, powered by \algname. Specifically, our planner comprises 15 Intelligent Driver Model (IDM)-based policies with varied parameters, in addition to one imitation policy. We utilize the same set of IDM policies as PDM and adopt the ego vehicle's prediction within \algname as the imitation policy. These policies are then simulated in parallel, interacting with \algname. This interaction is achieved by overriding the ego vehicle's next states with the output from the policy. Concurrently, the predictions of \algname for other agents are used as input for the next frame of the policy. After interacting for 4 seconds, we generate a series of rollouts for future states.

By evaluating these simulated rollouts, we can derive the expected return. Considering \algname operates stochastically, to make a more accurate estimation, we simulated and evaluated each policy for four times, then averaged these results to calculate the expected return.  Ultimately, we selected the output of the policy with the highest return as the next action to be taken.

In selecting our evaluation environment, we updated the original IDM-based environment by substituting IDM-controlled smart agents with those controlled by \algname. For a visual comparison of these two testing environments, one can refer to Appendix~\ref{sup: reactive simulation}. Compared to IDM, \algname-based smart agents exhibit more realistic and natural behavior, thus providing a more accurate representation of the planner's real-world performance. To ensure the accuracy of our assessment of the planner's performance, our experiments are conducted across both environments, with results presented side by side for reference. This method enables a comprehensive evaluation of our planning strategies in scenarios closely mimicking real-world driving conditions.



\subsection{Online Training}
\label{ape:C4}
We implement our experiments using an existing open-source RL framework \cite{Xu2021ALF}, which provides implementations of a number of standard RL algorithms. We use the Soft Actor-Critic (SAC) algorithm \cite{haarnoja2018soft} to train a simple policy model, which consists of a ResNet-18 model followed by a 2-layer MLP-based projection network, outputting a squashed Gaussian distribution representing the state-conditional action distribution. 

Using the same RL algorithm, network structure and hyperparameters, we train the policy network under the following two different environment setups respectively and compare the results:
\begin{itemize}
    \item \textbf{Setup 1: Log Playback}. Log playback environment is used in this setting which has no reactive capability.
    \item \textbf{Setup 2: World Model}. A learned world model is used to implement the environment, in which participating agents react according to ego vehicle's action.
\end{itemize}

 For each episode, the world model sees the current observation and unrolls one step conditioned on the policy model's output. we run 128 such environments in parallel. During the policy gradient update, we sample batches of 2048 state transitions from the replay buffer and optimize via Adam optimizer~\cite{kingma2014adam}.  Each policy model is trained for 40k steps. The learning rate is set to 1e-4 and is reduced to 2e-5 after 70\% of the training process. The training rewards can be found under the metrics section in Appendix.~\ref{sup: rl-metric}. During training, 20\% of the samples are used in an open-loop fashion, producing imitation losses. Empirically, we found that mixing open-loop samples during training helps speed up the training process. We then simulate each trained policy model under these environments and report metrics.
 
\section{Training setup}
\label{ape:D}

\subsection{Raster Input}
\label{ape:D1}
The raster layers encode various static elements and high definition map, which contrains:

\begin{itemize}
\item \textbf{Roadmap Raster} Represents the road network layout, including lanes, intersections, and other road features.
\item \textbf{Baseline Paths Raster} Encodes baseline paths within the road network of the scene.
\item \textbf{Route Raster}  Represents the ego vehicle's desired route or path.
\item \textbf{Drivable Area Raster} Represents areas considered drivable or navigable for vehicles within the scene.
\item \textbf{Speed Limit Raster}  Encodes speed limits at various locations within the scene.
\item \textbf{Static Agents Raster}  Represents positions of static agents (e.g., traffic cones) within the scene.
\item \textbf{Traffic Light Raster}  Encodes traffic light location and orientation at intersections within the scene.
\end{itemize}

\subsection{Data Augmentation}
\label{ape:D2}
To enhance the generalization capability of our model, we implemented various data augmentation techniques. First, we applied an agents and frame dropout strategy, randomly dropping certain agent tokens or the tokens of an entire frame with a probability of 0.1. Second, we employed a random crop strategy, randomly selecting segments from the entire scenario sequence based on a fixed window size for training. Additionally, we uniformly sampled rotation angles from \([- \frac{\pi}{2}, \frac{\pi}{2}]\) to rotate the entire scene and translated the whole scene in both x and y directions within a range of \([-5, 5]\) meters. Finally, beyond using the self-driving car as the central focus of the scene, we also randomly selected several agents of interest from the dataset to serve as the center of the scene coordinate system.

\subsection{Loss}
\label{ape:D3}
The loss function of \algname is composed of three main components: reconstruction loss, cross entropy loss, and multipath loss~\cite{chai2020multipath}. 

The reconstruction loss employs the L1 loss to supervise the image reconstruction part of the image autoencoder, which is defined as follows:

\begin{equation}
L_{\text{rec}} = \frac{1}{N_x N_y}\sum_{i,j} |I(i,j) - \hat{I}(i,j)|
\end{equation}

where \(N_x, N_y\) represent the number of 2D pixels, \(I\) is the image pixel, and \(\hat{I}\) is the reconstructed image. 

The cross entropy loss (CE) is responsible for evaluating the accuracy of categorical predictions made by the model. It is applied to both key tokens, such as ID, class and special tokens, as well as value tokens such as position (\(x, y\)), orientation (\(\theta\)), and dimensions (\(v_x, v_y, w, l\)). This loss is formulated as:
\begin{equation}
L_{\text{CE}} = \frac{1}{N_{k}} \sum_{k\in \{ID,class,special\}} \mathcal{H}(\hat{k}, k) + \frac{1}{N_{v}} \sum_{v\in \{ x,y,\theta, v_x, v_y, w, l\}} \mathcal{H}(\hat{v}, v) 
\end{equation}
where \(\mathcal{H}\) denotes the cross entropy, \(N_k\) and \(N_v\) are the counts of key and value tokens, respectively, and \(\hat{k}\) and \(\hat{v}\) are the predicted logits.

To enhance the model's motion prediction capability, we introduced trajectory prediction as an auxiliary task. The specific results and metric comparisons can be found in Appendix~\ref{sup: trajectory}. We employed the commonly used multipath loss~\cite{chai2020multipath}, which combines a classification loss (\(L_{cls}\)) with a minimum Average Displacement Error (\(L_{minADE}\)). The classification loss supervises the likelihood of different future paths, while the minADE measures the accuracy of the predicted trajectory (\(\hat{s}\)) against the ground truth (\(s\)). The balance between these two components is regulated by coefficients \(\alpha\) and \(\beta\):

\begin{equation}
L_{\text{traj}}^{i} = \alpha L_{cls}^{i} + \beta L_{minADE}(\hat{s}_{i}, s)
\end{equation}

In MCT, trajectory predictions are made at intervals of every \(n\) layers, resulting in multiple sets of trajectory prediction outcomes. As indicated above, \(i\) represents the results from different layers of trajectory prediction heads. 

Finally, the overall loss \(L\) is a weighted sum of these components, allowing for customized emphasis on different aspects of the model's predictions:
\begin{equation}
    L = \omega_{rec} * L_{\text{rec}} + \omega_{CE} * L_{CE} + \sum_{i} \omega_{traj}^i * L_{traj}^{i}
\end{equation}

where \(\omega_{rec}\), \(\omega_{CE}\), and \(\omega_{traj}^i\) are the weights applied to the reconstruction, cross entropy, and trajectory prediction losses, respectively. 

\subsection{Hyperparameters}
\label{ape:D4}
\subsubsection{Model configuration}
As shown in Table~\ref{tab:modelconfig}, we list out the detailed hyperparamters of three different model variants.

\begin{table}[ht]
\centering
\begin{tabular}{>{\raggedright}p{4cm}|>{\centering}p{2.5cm}>{\centering}p{2.5cm}>{\centering\arraybackslash}p{2.5cm}} 
\toprule
Hyperparameter & \algname-small & \algname-base & \algname-medium \\ 
\midrule
vocabulary size & \multicolumn{3}{c}{2972}  \\
block size & \multicolumn{3}{c}{2048} \\
meshgrid $[x, y, \theta, w, l, v_x, v_y ]$ & \multicolumn{3}{c}{$[ 0.2m, 0.2m, \frac{\pi}{100}, 0.5m, 0.5m, 0.25m/s, 0.25m/s]$}  \\
range $[x, y, \theta, w, l]$ & \multicolumn{3}{c}{$[(-100, 100)m, (-100, 100)m, (-\pi, \pi), (0, 7)m, (0, 15)m]$} \\
range $[v_x, v_y ]$ & \multicolumn{3}{c}{$[(0,25)m/s, (0,25)m/s]$ } \\
embedding dim & 384& 768 & 1024 \\
MCT backbone & gpt2-small & gpt2-base & gpt2-medium \\
$N_{heads}$ & 6 & 12 & 16 \\
$N_{layer}$ & 12 & 12& 24 \\
\# GRU decoder layer & 2 & 2 & 2 \\
chunking size $T$(sec) & 2 & 2 & 2\\
visual encoder & ResNet18 & ResNet34 & ResNet50 \\
total params & 55.8M &184M & 523M \\ 
\bottomrule
\end{tabular}
\caption{Model configurations of \algname.}
\label{tab:modelconfig}
\vspace{-8mm}
\end{table}

\subsubsection{Training configuration}
We detail all training specifics in Table~\ref{tab:traininingdetails}. Our models undergo an initial training phase of 10 epochs on the nuPlan Dataset, followed by a fine-tuning stage on the Waymo Open Dataset for an additional 10 epochs. The \algname-medium model is trained using 8 A100 GPUs with a period of 3 days.
\begin{table}[ht]
\centering
\begin{tabular}{>{\raggedright\arraybackslash}p{4cm}|>{\centering\arraybackslash}p{2.5cm}} 
\toprule
Hyperparameter & Value \\
\midrule
lr &  2e-4 \\
weight decay & 1e-3 \\
optimizer & adamw~\cite{loshchilov2018decoupled} \\
lr scheduler & multistep lr \\ 
batchsize & 64 \\
accumulate grad batches & 2\\
training epoch & 10 \\
decay milestones & [6, 8] \\
decay rate & 0.1 \\
precision & fp16 \\
\hline
$\omega_{rec}$ & 1.0 \\
$\omega_{CE}$ & 1.0 \\
$\omega_{traj}^{0}$ & 0.25 \\
$\omega_{traj}^{1}$ & 0.25 \\
$\omega_{traj}^{2}$ & 0.5 \\
$\omega_{traj}^{3}$ & 0.5 \\
\hline
temperature & 1.1 \\
topk & 40 \\
temperature at SceneGen & 1.25 \\
topk at SceneGen & 400 \\
condition length & 3 \\
decay rate $\gamma$ & 1.2 \\
chunking time horizon $T$ & 2s \\

\bottomrule
\end{tabular}
\caption{Training hyperparameters of \algname.}
\vspace{-8mm}
\label{tab:traininingdetails}
\end{table}


\section{Ablation Study}
\label{sec: ablation study}

\subsection{Effectiveness of NAR conversion}
\label{sup: nar ablation}
To accelerate the model's inference process, we employed Non-Autoregressive (NAR) conversion methods described in Section~\ref{sec:convert}, transforming the full autoregressive (full-AR) model into a partially autoregressive (partial-AR) model. The specific experimental results, as outlined in Table~\ref{tbl:nar_conversion}, reveal that the transition to a partial-AR model results in a significant acceleration ($\times 132$ speed up) without any noticeable degradation across various simulation realism metrics. This transformation demonstrates the efficacy of NAR conversion methods in enhancing computational efficiency while maintaining the quality of simulation outcomes.

\begin{table*}[h]
\begin{center}
    \resizebox{0.8\textwidth}{!}{%
        \begin{tabular}{c|c|cc|ccc}
        \hline
    \multicolumn{1}{c|}{Inference Mode} & \multicolumn{1}{c|}{Inference Speed (FPS)}& \multicolumn{1}{c}{Meta Metric$\uparrow$} & \multirow{2}{*}{minADE$\downarrow$} &
    \multicolumn{1}{c}{Kinematic Metrics$\uparrow$}&
    \multicolumn{1}{c}{Interactive Metrics$\uparrow$}&
    \multicolumn{1}{c}{Map-based Metrics$\uparrow$}\\
    \multicolumn{1}{c|}{}  & \multicolumn{1}{c|}{} & \multicolumn{1}{c}{} &
    &
    \multicolumn{1}{c}{}&
    \multicolumn{1}{c}{}&
    \multicolumn{1}{c}{}\\
\hline 
full-AR & 0.69 & 0.645 & 1.548 & 0.4022 & 0.7651 & 0.8334\\
partial-AR & \textbf{82}($\times 132$) & 0.646 & 1.584 & 0.4036 & 0.7662 & 0.8331 \\
        \hline
        \end{tabular}
        }
\end{center}
\caption{Comparison of full-AR and partial-AR inference mode. The inference speed is measured with a single A100 GPU. The transition from full-AR to partial-AR mode results in a significant acceleration ($\times 132$ speed up) without any noticeable degradation across various simulation realism metrics.}
\label{tbl:nar_conversion}
\end{table*}

\subsection{Effectiveness of Prediction Chunking and Temperal Aggregation.}
\label{ape:E2}
As shown in Table~\ref{tbl:chunking_ablation}, we present the ablation study results on the Waymo Sim Agents validation set. It is evident that incorporating prediction chunking as an auxiliary task alone leads to a 1\% enhancement in the realism meta-metric, with particularly notable improvements observed in the interactive and map-based metrics. Moreover, the minimum Average Displacement Error (minADE) significantly decreases. Further enhancements are attained through temporal aggregation, where the ensemble of multiple predictions from chunked results yields an additional 1.2\% improvement in the meta-metric. It is worth mentioning that the kinematic metrics are the primary contributors to this overall improvement. These experiments underscore the effectiveness of the Prediction Chunking and Temporal Aggregation modules in increasing the realism of simulations conducted by \algname.

\label{sup:chunk}

\begin{table*}[h]
\begin{center}
    \resizebox{0.8\textwidth}{!}{%
        \begin{tabular}{cc|cc|ccc}
        \hline
    \multicolumn{1}{c}{Prediction} & \multicolumn{1}{c|}{Temporal} & \multicolumn{1}{c}{Meta Metric$\uparrow$} & \multirow{2}{*}{minADE$\downarrow$} &
    \multicolumn{1}{c}{Kinematic Metrics$\uparrow$} &
    \multicolumn{1}{c}{Interactive Metrics$\uparrow$} &
    \multicolumn{1}{c}{Map-based Metrics$\uparrow$} \\
    \multicolumn{1}{c}{Chunking} & \multicolumn{1}{c|}{Aggregation} & \multicolumn{1}{c}{} &
    &
    \multicolumn{1}{c}{} &
    \multicolumn{1}{c}{} &
    \multicolumn{1}{c}{} \\
\hline 
& & 0.624 & 1.626 & 0.3903 & 0.7508 & 0.7903 \\
\checkmark & & 0.634 & \textbf{1.562} & 0.3700 & \textbf{0.7674} & 0.8299 \\
\checkmark & \checkmark & \textbf{0.646} & 1.584 & \textbf{0.4036} & 0.7662 & \textbf{0.8331} \\
        \hline
        \end{tabular}
    }
\end{center}
\caption{Ablation study of prediction chunking in Waymo Sim Agents benchmark. These experiments demonstrate the efficacy of Prediction Chunking and Temporal Aggregation modules in enhancing the realism of simulations conducted by \algname.}
\label{tbl:chunking_ablation}
\end{table*}

\subsection{Ablation Study of Decay Rate $\gamma$}
\label{ape:E3}
\begin{figure}[h]
\centering
\includegraphics[width=1.0\textwidth]{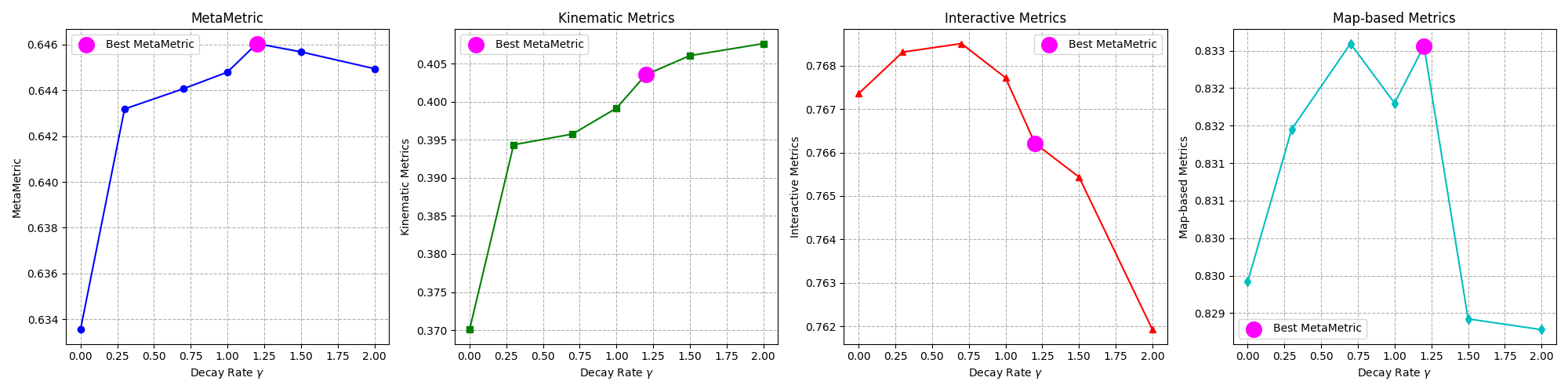}
\caption{Impact of Decay Rate \(\gamma\) on the Realism Metrics in the Waymo Sim Agents Benchmark. This figure illustrates the variation curves of the overall MetaMetric, Kinematic metrics, Interactive metrics, and Map-based metrics as a function of decay rate \(\gamma\), from left to right. The experimental results highlight that the overall meta-metric reaches its peak performance at \(\gamma = 1.2\), which is highlighted with magenta.}
\label{fig:decay_vs_metascore}
\end{figure}

In this section, we explore the optimal setting for the decay rate \(\gamma\) in our temporal aggregation process, as depicted in Figure~\ref{fig:decay_vs_metascore}. Setting \(\gamma = 0.0\) equates to omitting temporal aggregation entirely. As we increase \(\gamma\), future predictions within the chunking data are progressively weighted more heavily. Observing the components of the meta-metric, it becomes evident that a higher decay rate $\gamma$ distinctly improves the kinematic metrics while leading to a decrease in the interactive and map-based metrics. The overall meta-metric peaks at \(\gamma = 1.2\). This suggests that by incorporating predictions that extend further into the future, agents are better able to adhere to their initial goals, thereby achieving higher kinematic metrics. However, this reliance on more distant predictions in chunked data tends to overlook interactions between agents, and between agents and the map at future time points, resulting in reduced interactive and map-compliance metrics.

\subsection{Ablation Study of Temperature}
\label{ape:E4}
\begin{figure}[h]
\centering
\includegraphics[width=1.0\textwidth]{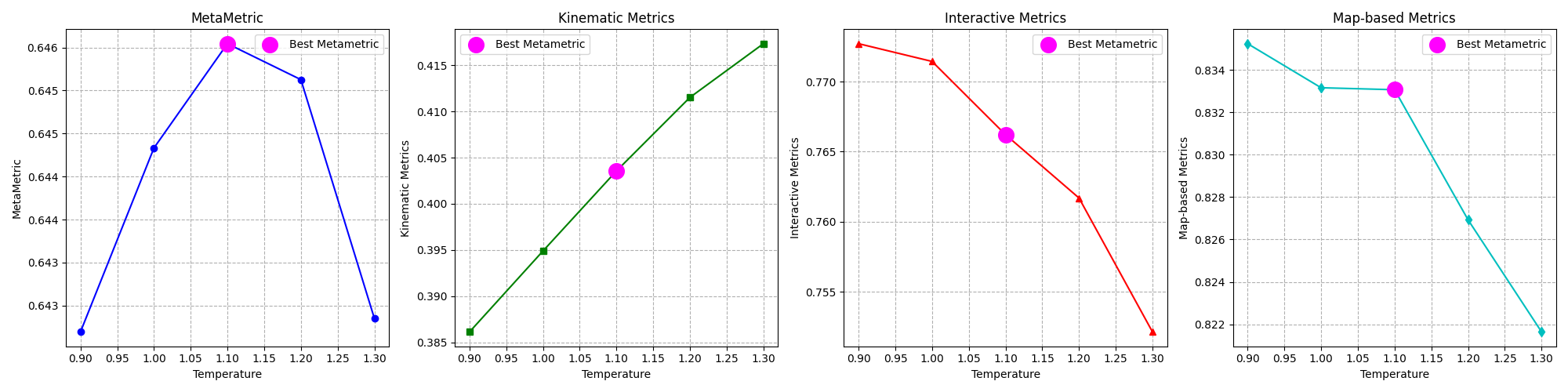}
\caption{The graph illustrates the relationship between temperature settings and Realism Metrics in the Waymo Sim Agents Benchmark. It highlights how varying temperature levels affect overall simulation performance, with a focus on the interplay between kinematic metrics, interactive behaviors, and map compliance. The chart specifically showcases the optimal meta-metric peak at a temperature of 1.1, while also detailing the divergent trends of increasing kinematic metrics and decreasing interactive and map-based metrics performance at higher temperatures.}
\label{fig:temp_vs_metascore}
\end{figure}

In this section, we conduct a detailed analysis of how varying the temperature parameter affects the performance and behavior of simulations, as illustrated in Table~\ref{fig:temp_vs_metascore}. The data clearly shows that the meta-metric peaks at a temperature of 1.1. Furthermore, as the temperature increases, kinematic metrics show a monotonically increasing trend, while both interactive and map-based metrics exhibit a monotonic decrease. This pattern indicates that higher temperatures facilitate better mode coverage by enabling more frequent sampling of low probabilistic states. However, such states may lead to an increase in collisions and deteriorate map compliance. Visual observations reveal that higher temperatures result in more aggressive behaviors among road participants, such as increased collisions or jaywalking. Conversely, lower temperatures are associated with more conservative behaviors, including more obedient drivers and more cautious pedestrians. This parameter thus offers a lever to more precisely control the behavior of traffic participants, enabling effective evaluation of driving policies across diverse scenarios.

\subsection{Ablation Study of the Number of Conditioned Frames}
\label{ape:E5}
\begin{figure}[h]
\centering
\includegraphics[width=1.0\textwidth]{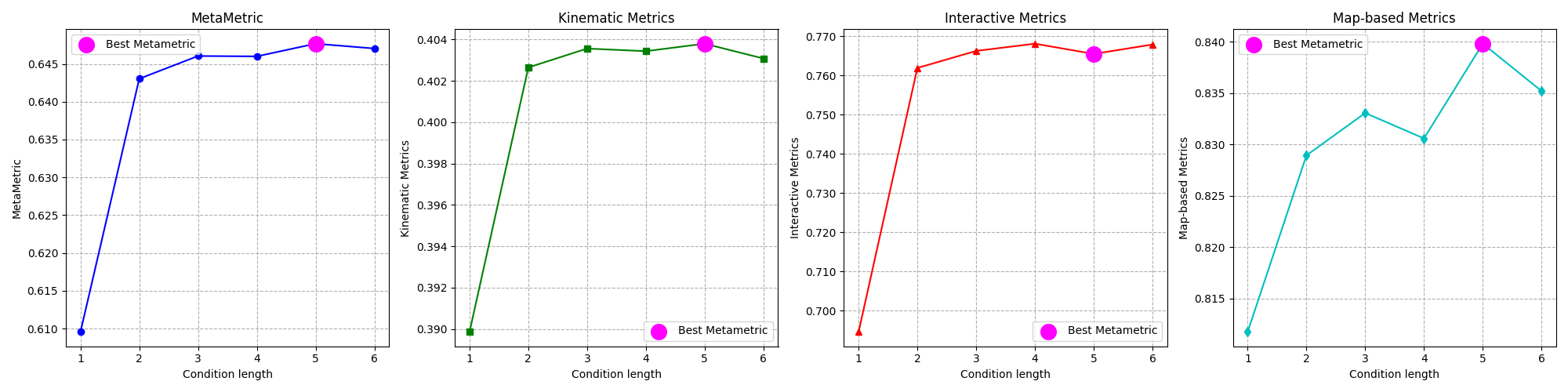}
\caption{Impact of the number of conditioned frames on the Realism Metrics in the Waymo Sim Agents Benchmark.}
\label{fig:cond_vs_metascore}
\end{figure}

In this section, we explore the effects of varying the length of conditioned history frames within the autoregressive process. As depicted in Table~\ref{fig:cond_vs_metascore}, the performance is notably poor when there is a lack of historical input (condition length = 1, i.e only current frame). With the increment in input frame length, the overall meta-metric reaches its peak at 5 frames, corresponding to a 2-second history, after which the improvement plateaus. This suggests that for the purpose of simulation, a 2-second window of information is sufficiently informative. However, considering that increasing the conditioned length significantly adds to computational demands, thereby affecting the efficiency of downstream tasks, we opt for a conditioned frames length of 3 (1-second history). This decision balances the need for sufficient historical context with the imperative of maintaining computational efficiency.

\section{Qualitative Analysis}
\label{sup: quanlitative}
In this section, we will analyze the visual analysis of \algname under different tasks. The visualization reflects \algname's excellent capability in controlled scene generation, its realistic simulation of diverse complex driving scenarios, and provides a visual comparison with traditional IDM agents. This fully demonstrates the superiority of \algname as a controllable realistic data-driven simulator.

\subsection{Scene Generation}
\label{ape:F1}
\begin{figure*}[ht]
    \centering
    \begin{subfigure}{\textwidth}
        \centering
        \includegraphics[width=0.9\textwidth]{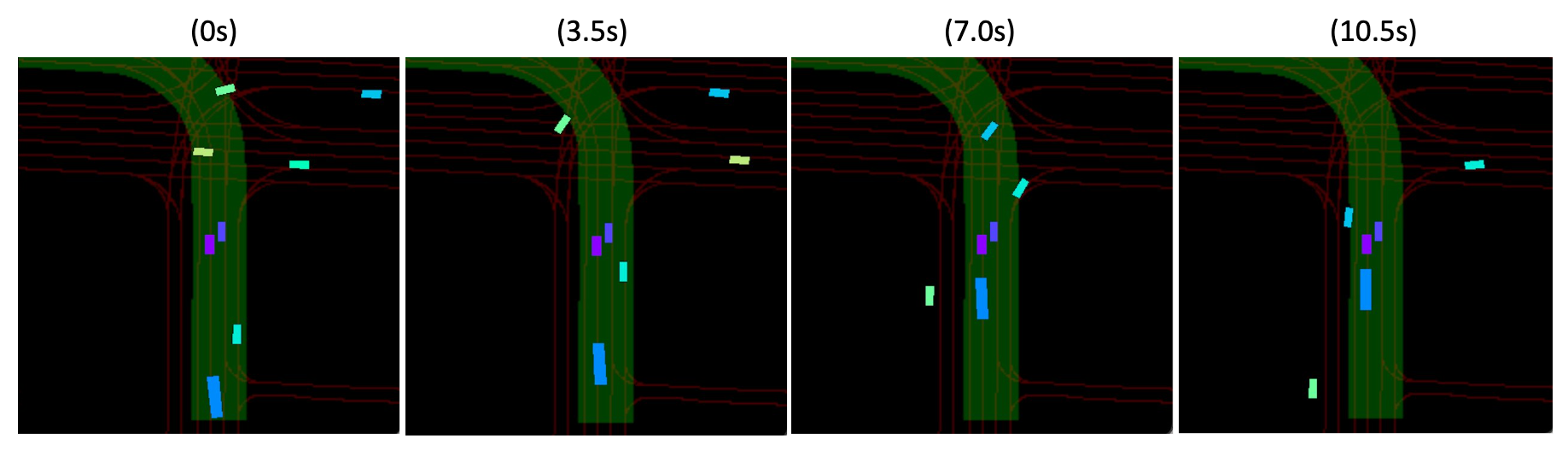}
        \subcaption{scenario prompt: low magnitude speed}
    \end{subfigure}
    \\
    \begin{subfigure}{\textwidth}
        \centering
        \includegraphics[width=0.9\textwidth]{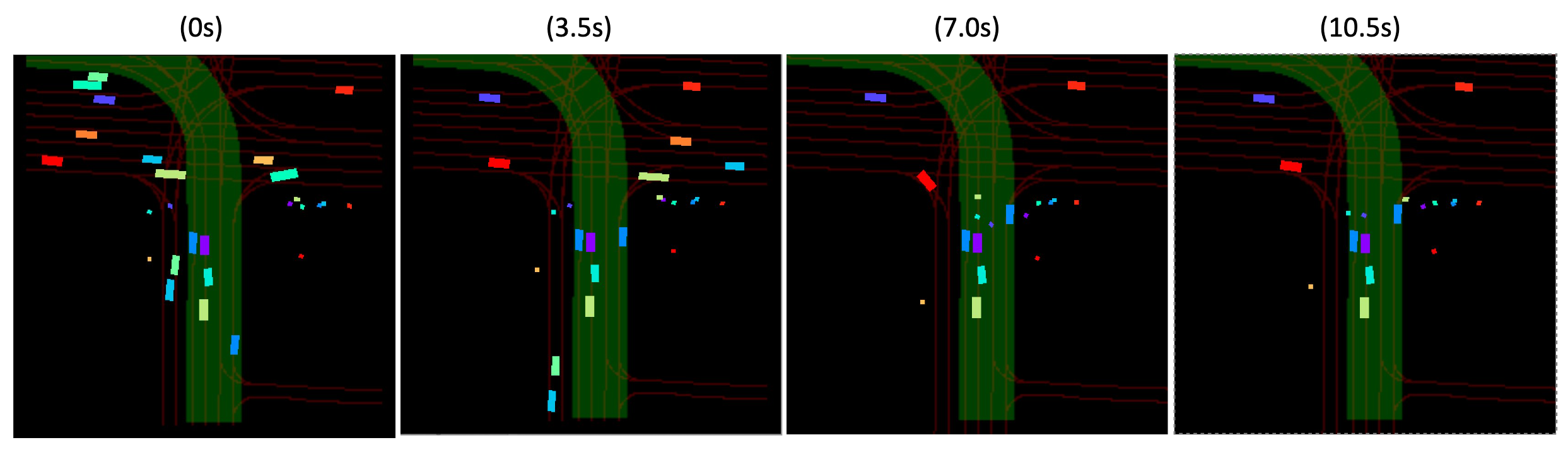}
        \subcaption{scenario prompt: waiting for pedestrian to cross}
    \end{subfigure}
    \\  
    \begin{subfigure}{\textwidth}
        \centering
        \includegraphics[width=0.9\textwidth]{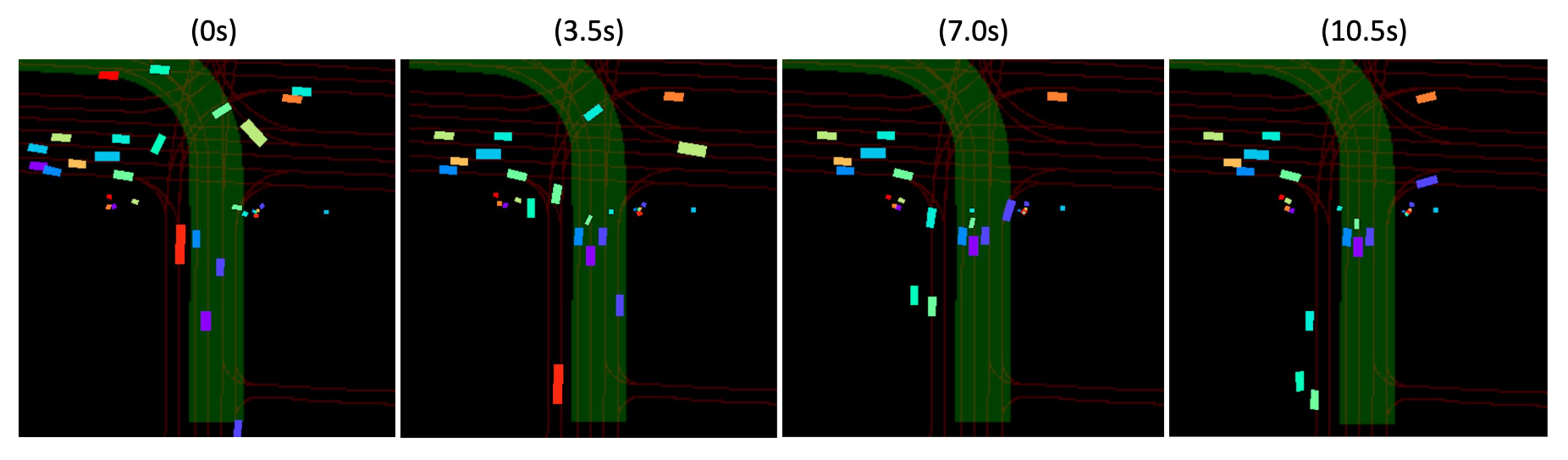}
        \subcaption{scenario prompt: behind bike}
    \end{subfigure}
    \\        
    \caption{We created and simulated three distinct scenarios using the same static map but with different scenario descriptions, including the scenario prompts and the classes and numbers of agents involved. It is worth noting that the 0th frame in the image is generated through fully autoregressive scene generation, while subsequent simulations are generated through partial-autoregressive scene extrapolation.}
    \label{fig:scene generation}
\end{figure*}

Here we demonstrate visualizations of scene generation based on map and scenario description prompts. By providing a static map, specifying the types and numbers of agents, and offering scenario description prompts, we can autoregressively generate corresponding initial states and simulate future based on the initial states. As illustrated in Figure~\ref{fig:scene generation}, under the same map input, the creation of distinctly different scenes is facilitated by varying the scenario descriptions and the numbers of objects. The 0th frame in the image is generated through fully autoregressive scene generation, while subsequent simulations are generated through partial-autoregressive scene extrapolation. Examples include low magnitude speed, waiting for pedestrian to cross, or controlling the position of objects, such as behind bike. Through such controlled methods, we are able to generate a large number of driving scenarios, thereby mitigating the issue of data scarcity.

\subsection{Diverse Future}
\label{ape:F2}
\begin{figure*}[ht]
    \centering
    \begin{subfigure}{\textwidth}
        \centering
        \includegraphics[width=0.36\textwidth]{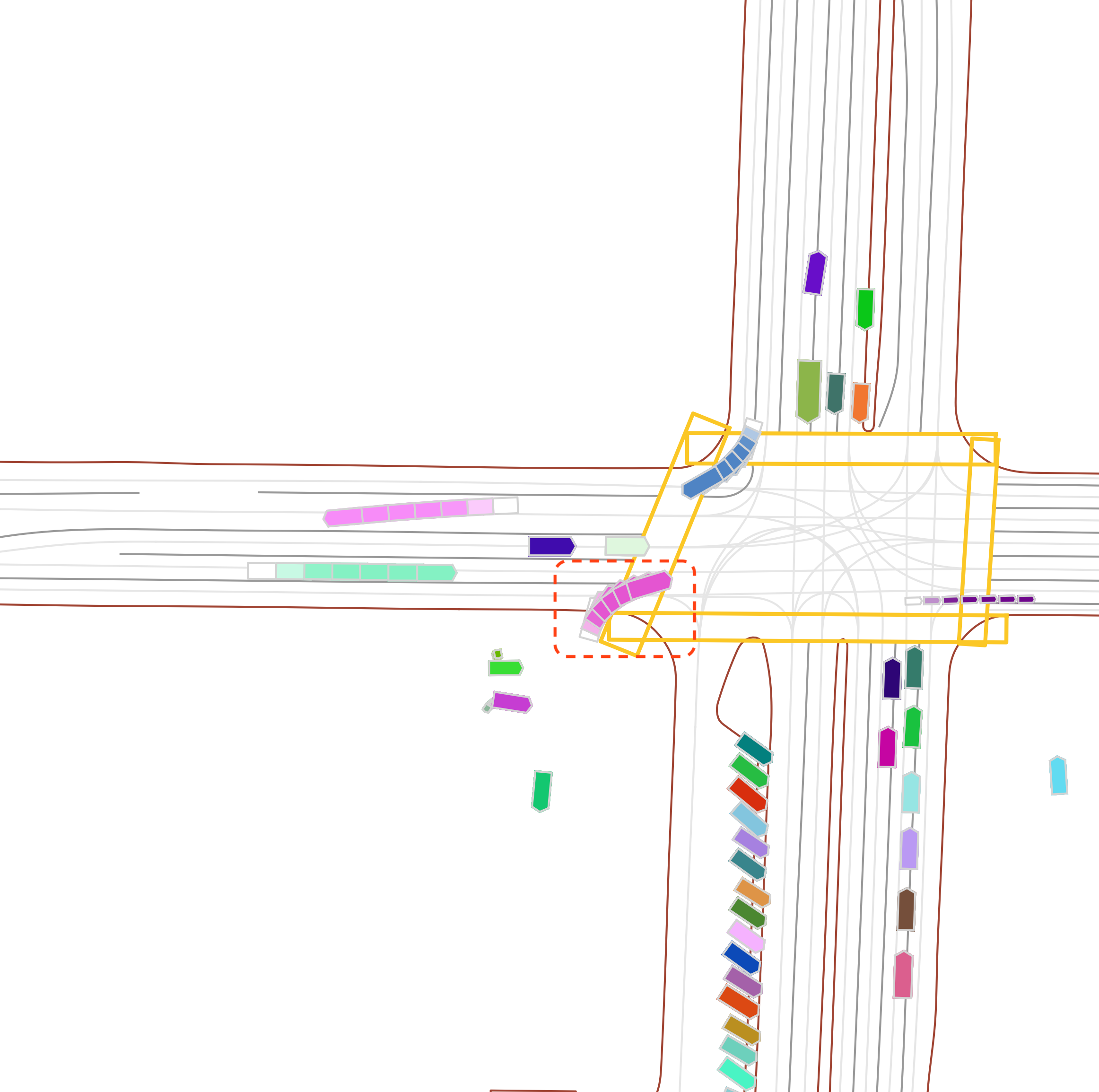}\hspace{0.5cm}
        \includegraphics[width=0.36\textwidth]{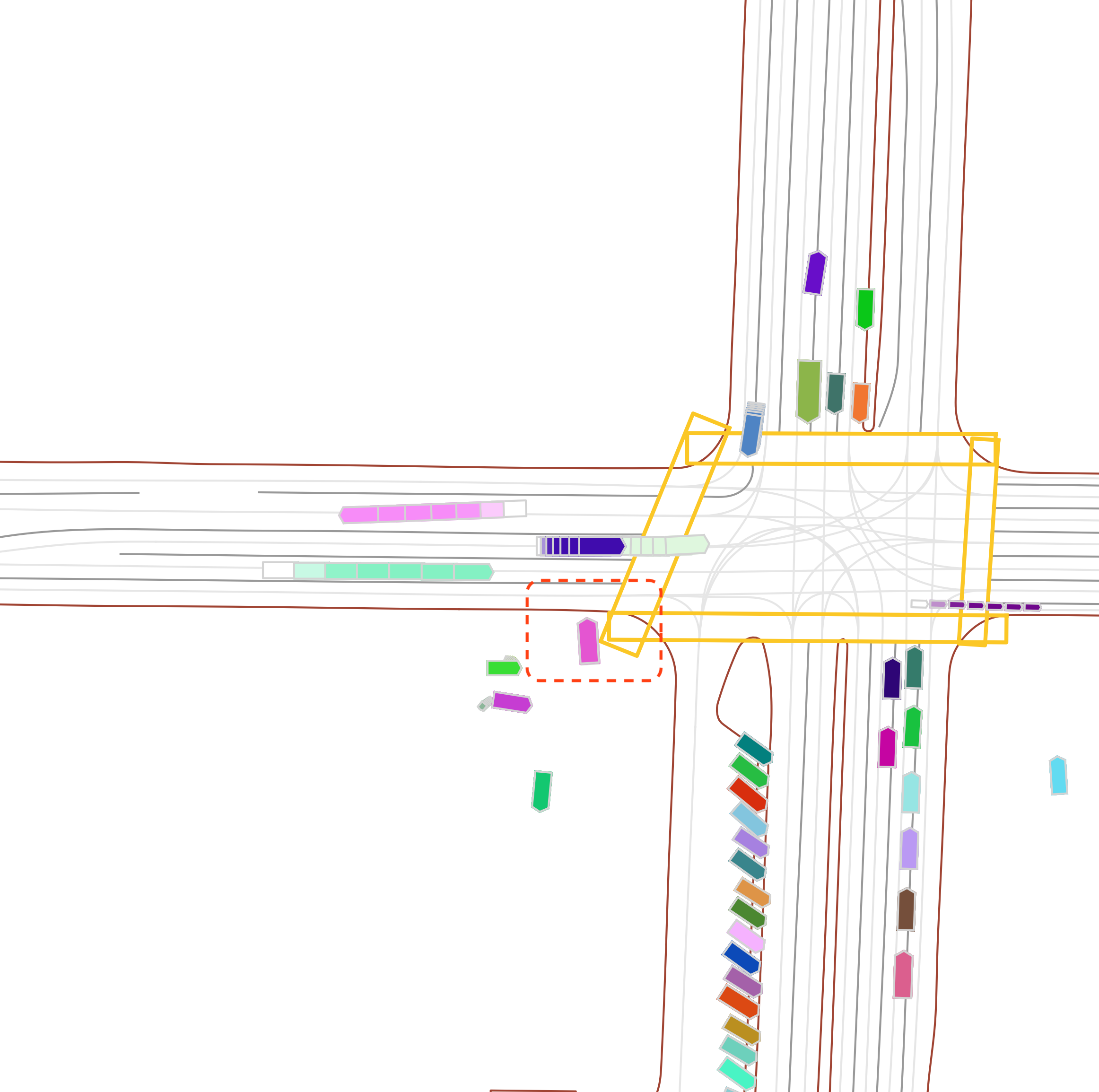}
        \subcaption{Left: proceed, Right: yield}
    \end{subfigure}
    \\
    \begin{subfigure}{\textwidth}
        \centering
        \includegraphics[width=0.36\textwidth]{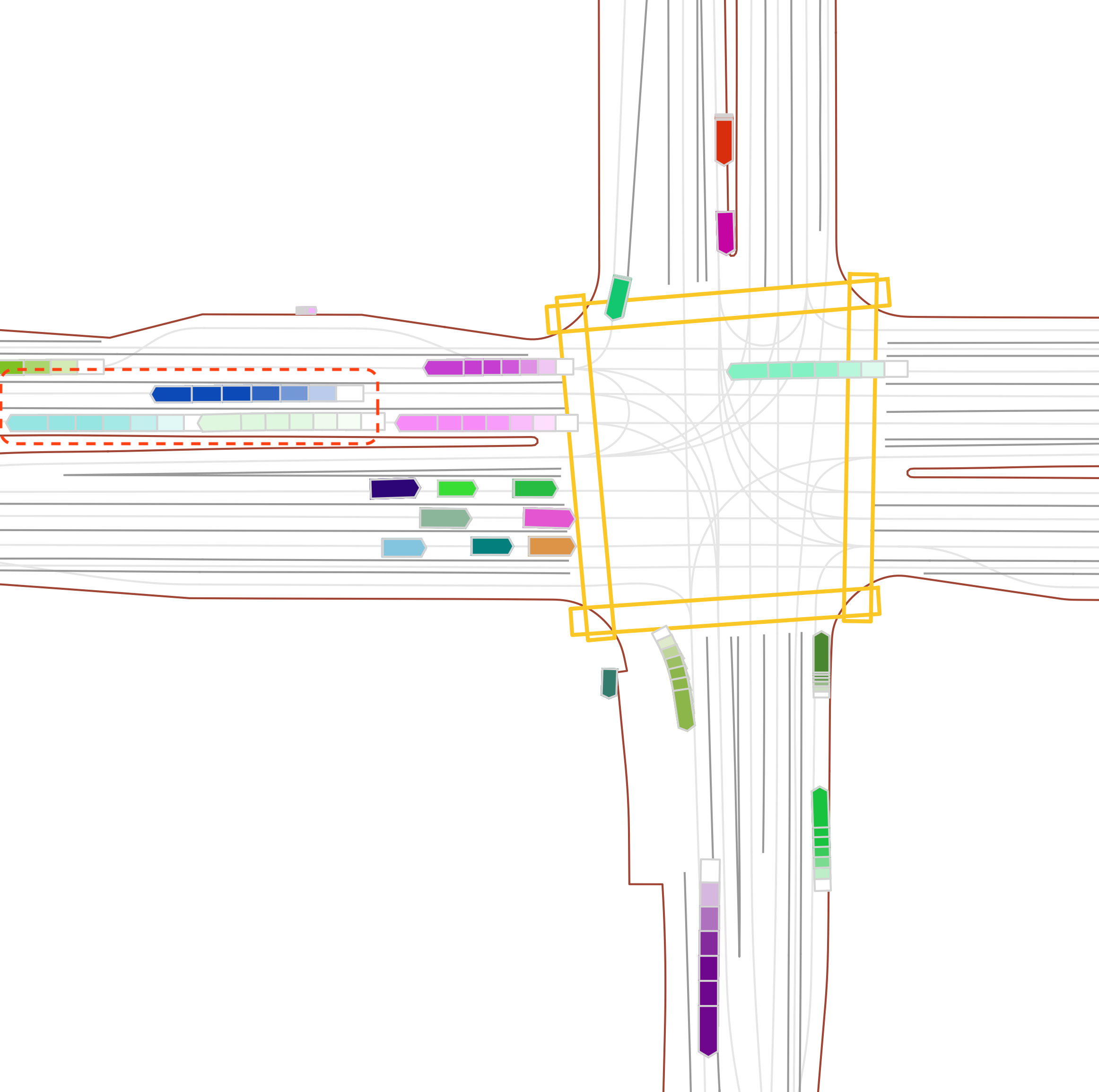}\hspace{0.5cm}
        \includegraphics[width=0.36\textwidth]{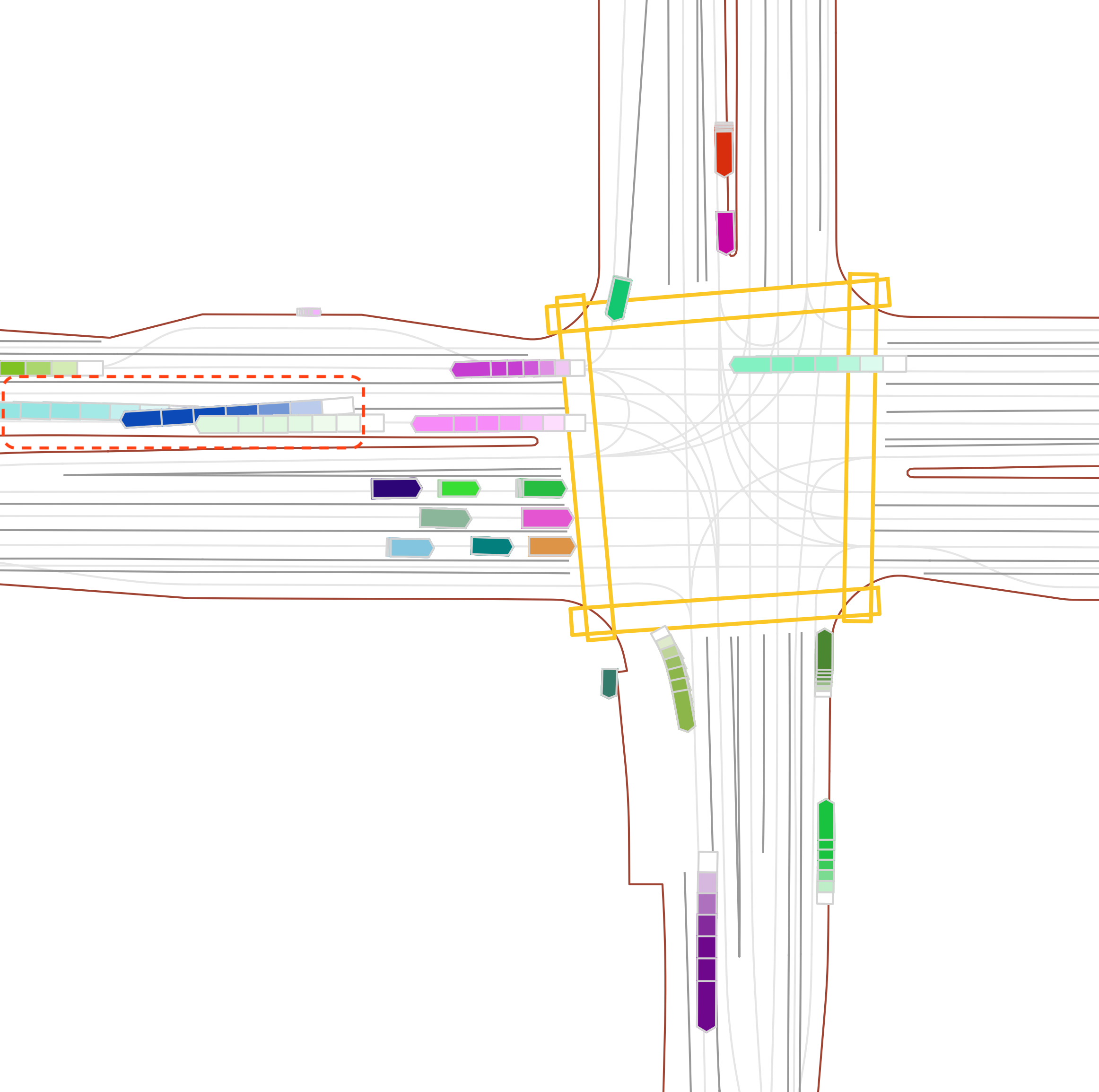}
        \subcaption{Left: maintain, Right: change lane}
    \end{subfigure}
    \\  
    \begin{subfigure}{\textwidth}
        \centering
        \includegraphics[width=0.36\textwidth]{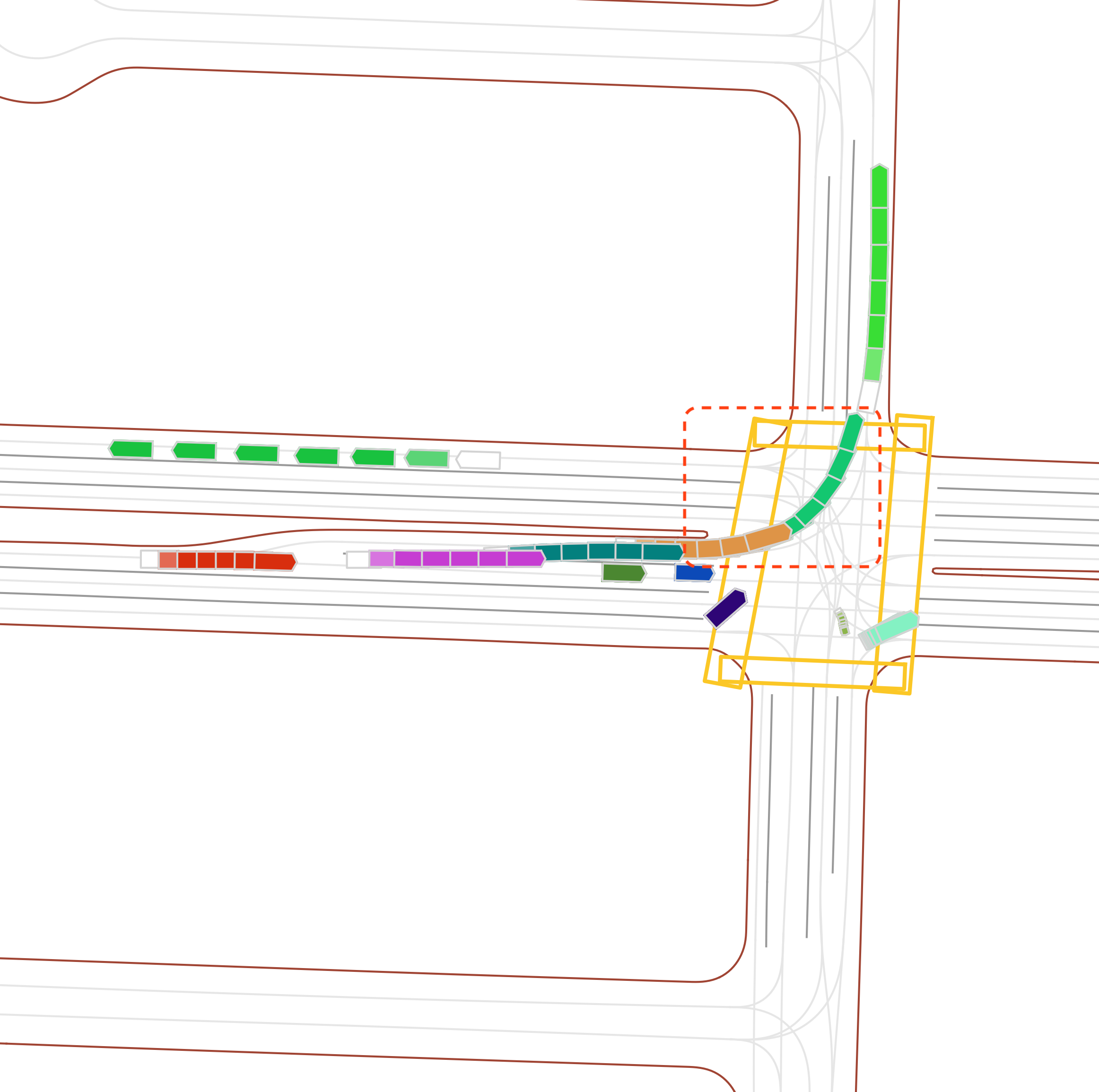}\hspace{0.5cm}
        \includegraphics[width=0.36\textwidth]{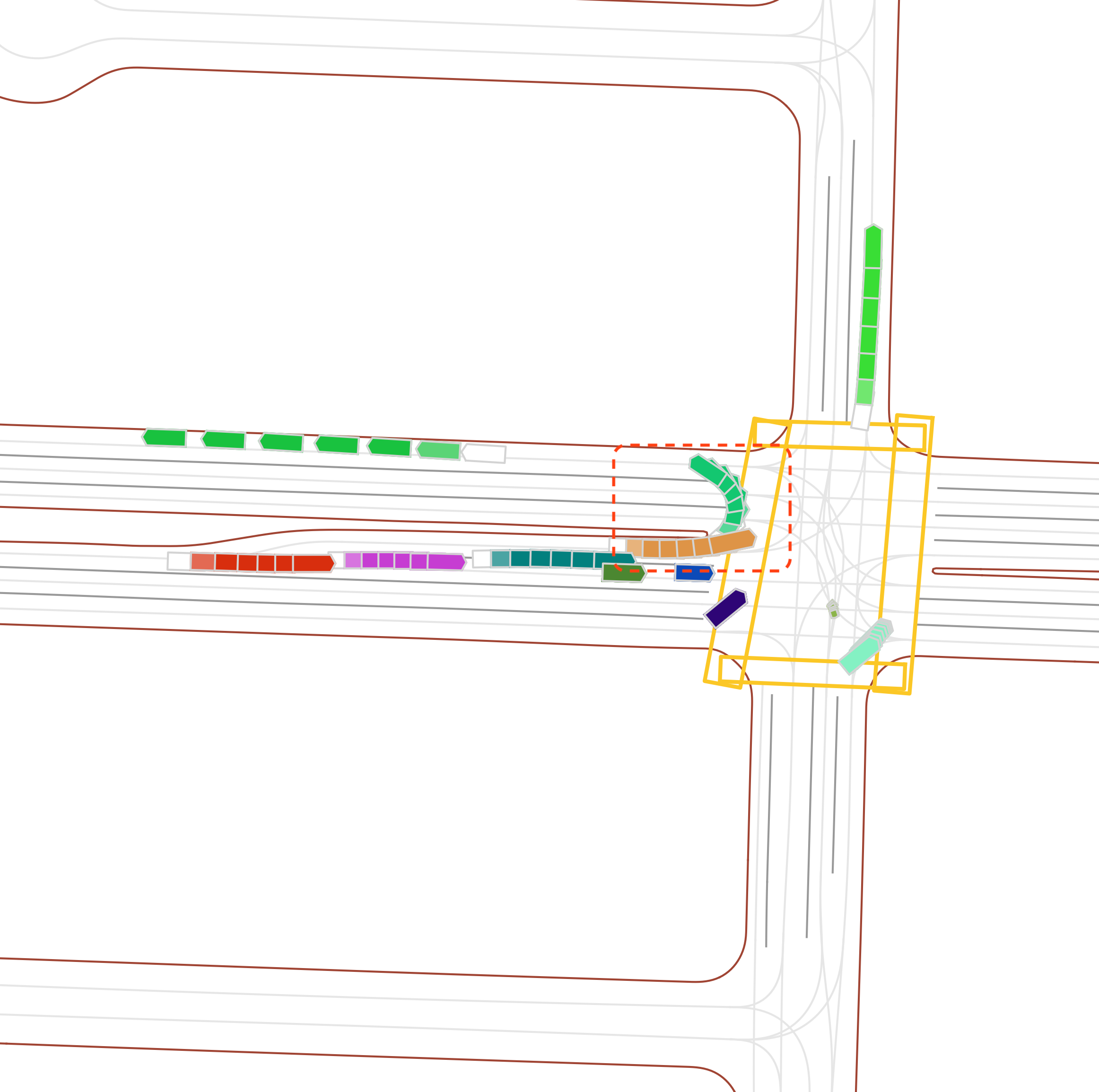}
        \subcaption{Left: left-turn, Right: u-turn}
    \end{subfigure}
    \\        
    \caption{The left and right images display the outcomes simulated under the same initial conditions. We overlap the 5 seconds future trajectories of all agents on a single image to represent the motion of each agent. The areas of interest in the image are highlighted with red dashed lines, which shows that simulated scenarios with identical initial conditions can lead to significantly different behaviors of the agents.}
    \label{fig:scenes}
\end{figure*}

\algname demonstrates exceptional efficacy in generating diverse, interactive and rational driving behaviors within identical map settings. Fig.~\ref{fig:scenes} show three pair of samples.

In Fig.~\ref{fig:scenes} (a), the red car proceeds to make a right turn because there is sufficient distance from the approaching green vehicle, demonstrating the model's effective distance assessment for safe maneuvers. Conversely, the right panel depicts the red car yielding to faster-moving green vehicle, illustrating the model's prioritization of safety in tighter traffic scenarios.

On the left side of Fig.~\ref{fig:scenes} (b), all vehicles stick to their predetermined routes. However, the right side illustrates a more complex scenario where a light blue car accelerates and changes lanes, forcing the dark blue car following it to also switch lanes in order to avoid congestion initiated by the leading vehicle.

Lastly, Fig.~\ref{fig:scenes} (c) illustrates the model's ability to choose highly probable actions, like making a left turn, as well as to perform less typical maneuvers, such as U-turns, given similar circumstances. This adaptability underscores the model's sophisticated decision-making capabilities across diverse traffic situations.

\subsection{Reactive Simulation}
\label{sup: reactive simulation}
\begin{figure*}[ht]
    \centering
    \begin{subfigure}{\textwidth}
        \centering
        \includegraphics[width=0.9\textwidth]{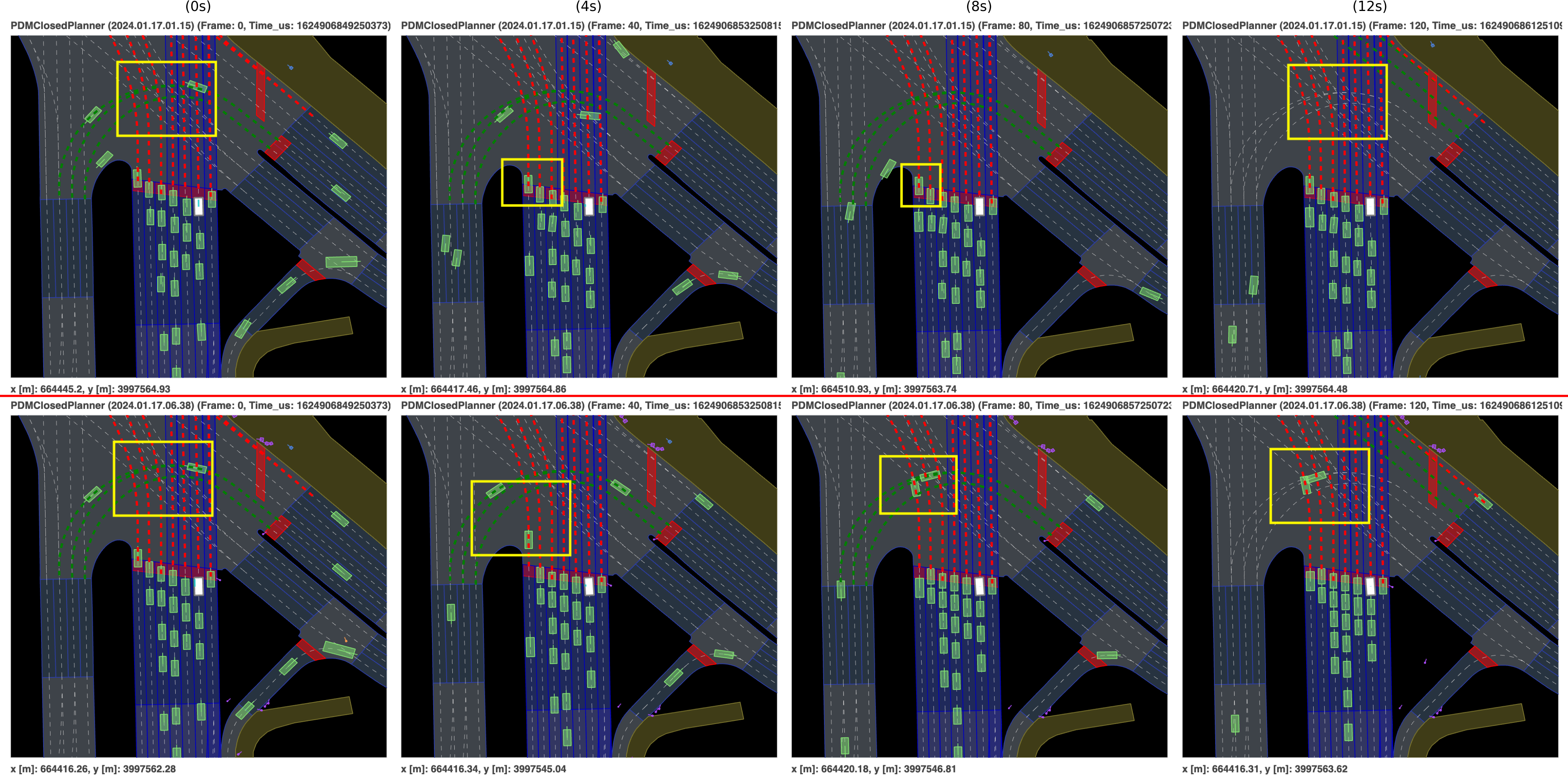}
        \subcaption{In the IDM environment, one vehicle is running through a red light and collides with another vehicle which is making a left turn.}
    \end{subfigure}
    \\
    \begin{subfigure}{\textwidth}
        \centering
        \includegraphics[width=0.9\textwidth]{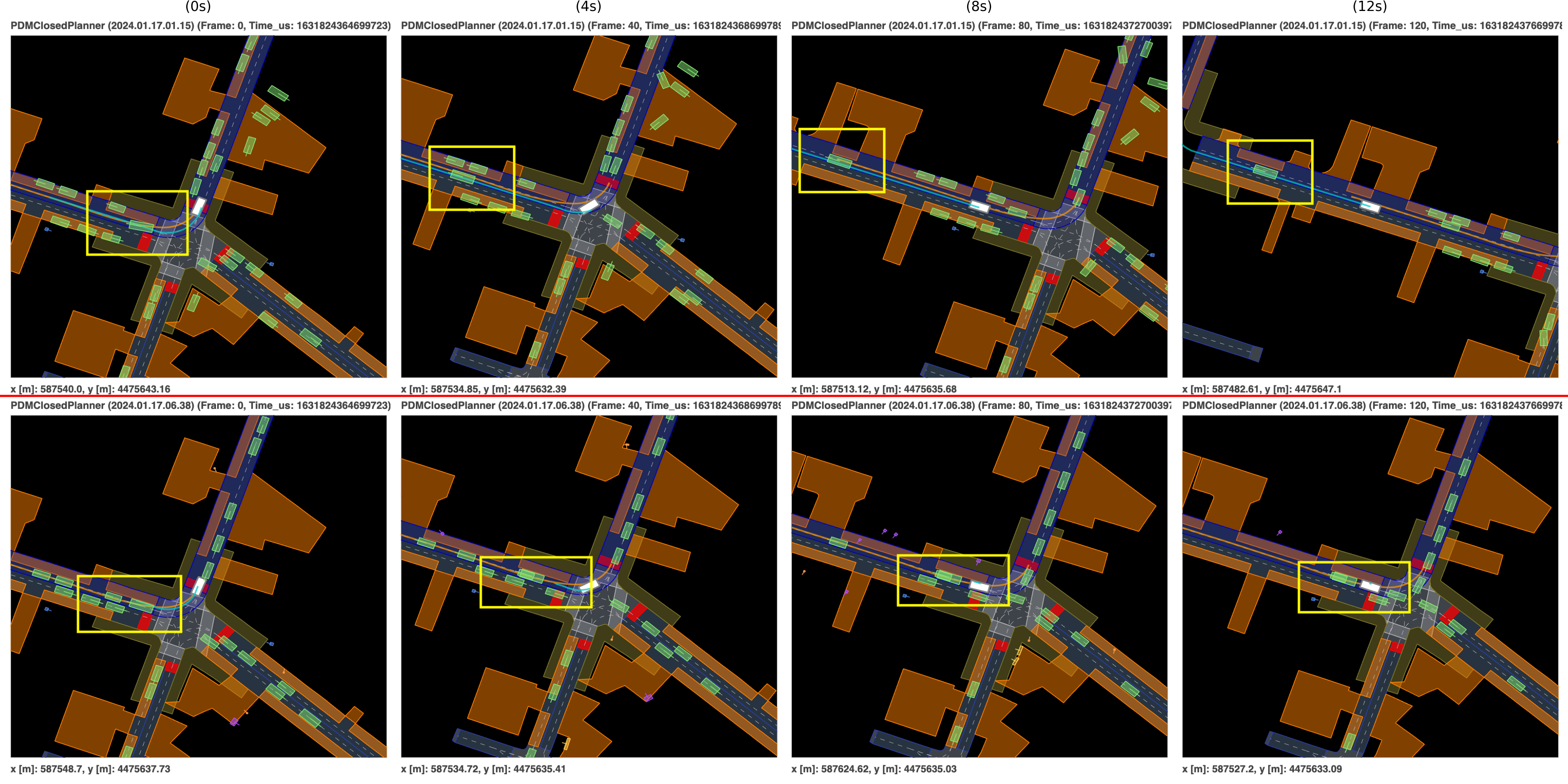}
        \subcaption{In the IDM environment, the bus collides with the vehicle in front of it.}
    \end{subfigure}
    \\ 
    \begin{subfigure}{\textwidth}
        \centering
        \includegraphics[width=0.9\textwidth]{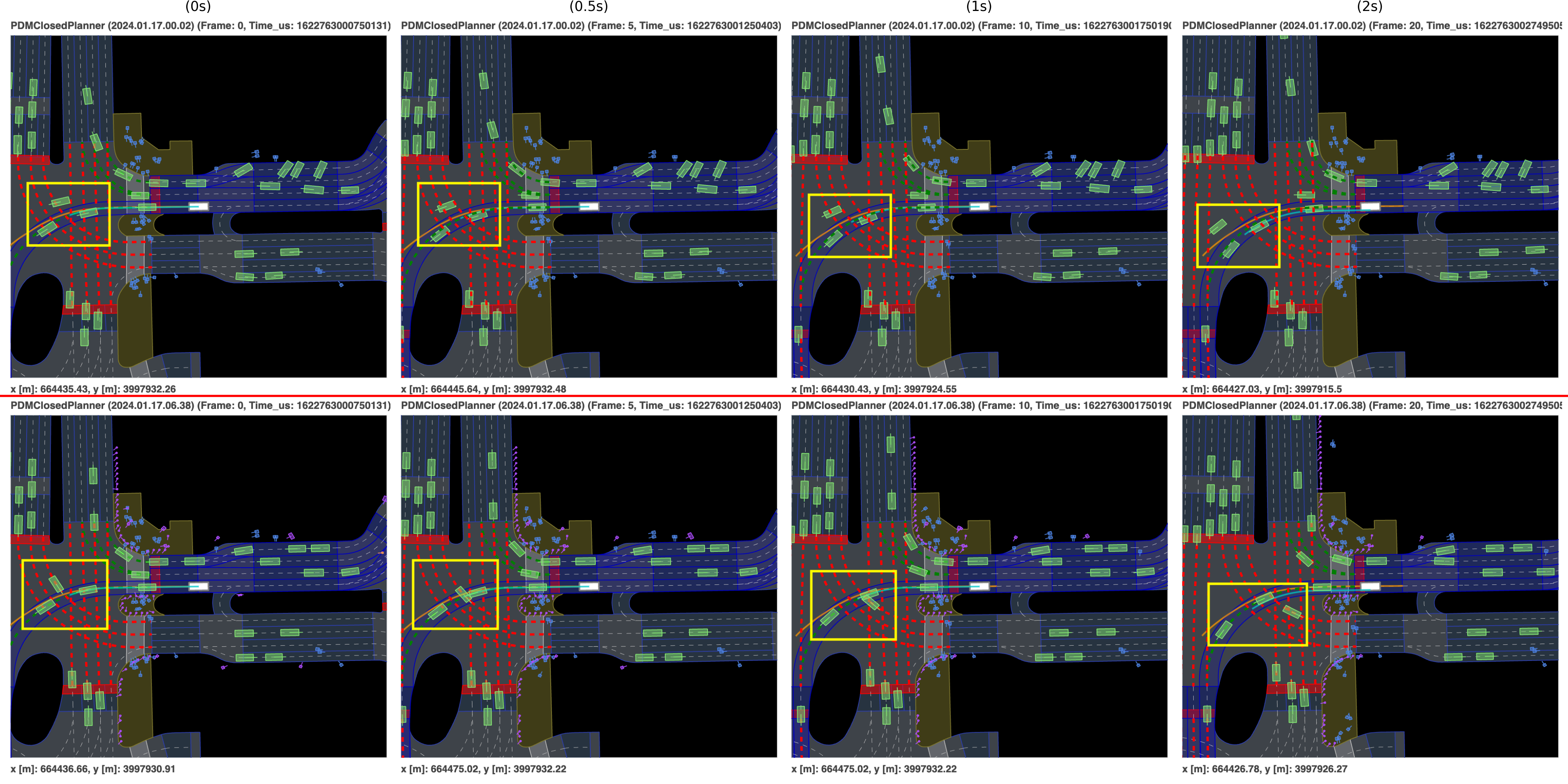}
        \subcaption{In the IDM environment, one vehicle is running a red light and collides with another left-turning vehicle.}
    \end{subfigure}
    \\  
    \caption{Agents' behaviours comparison under \algname's environment (first row) and IDM environment (second row). Interesting areas are highlighted with a yellow square.}
    \label{fig:wm_sim_vis}
\hfill
\end{figure*}

Fig.~\ref{fig:wm_sim_vis} illustrates the superior realism of \algname compared to the Rule-based reactive environments (IDM). The first row shows agents' movement generated by \algname and the second row is for IDM. These visualizations reveal that the simplistic IDM environment falls short of providing an adequate reactive simulation for planning policies. Conversely, \algname presents an efficient and more human-like alternative.

In Figure (a), we observe collisions between agents in the IDM environment, where a vehicle disregards traffic lights and collides with another vehicle making a left turn. In contrast, in \algname's environment, vehicles exhibit more human-like behavior, avoiding any collisions.

In the second row of Figure (b), a bus fails to stop adequately behind another vehicle, leading to a rear-end collision. Meanwhile, in the top row showcasing \algname's environment, the bus stops appropriately, avoiding any collision with the vehicle in front.

Finally, Figure (c) also highlights a scenario in the IDM environment where a vehicle ignores a traffic light and collides with another vehicle executing a left turn. However, in the environment simulated by \algname, vehicles navigate the intersection smoothly, without any collision.

\section{WOD Motion Results}
To further validate its capabilities as a foundation model for multiple tasks, we also benchmarked \algname's performance in motion prediction tasks. Utilizing the latent features corresponding to each agent, future 8 seconds trajectories and their associated probabilities are predicted through a simple multilayer perceptron (MLP) prediction head. Similar to the Waymo motion prediction benchmark, we adopted 6 distinct modes and supervised the model using a multipath loss.
\label{sup: trajectory}
\subsection{WODM Validation Set}
\label{ape:G1}
 As shown in Table~\ref{tab:WODmotion}, we compare the performance of \algname-medium on the WOD motion validation set against other models. It's important to note that, to ensure a fair comparison, all models selected for this analysis are end-to-end models, meaning they do not incorporate additional post-processing or ensembling in their results. Traditionally, agent-centric models have led the way in motion prediction due to their more unified and easier-to-learn representation forms, outpacing scene-centric models. Despite this, our findings reveal that \algname-medium significantly surpasses other scene-centric models, such as the SceneTransformer~\cite{ngiam2021scene}, and even approaches the performance of agent-centric models like MTR-e2e~\cite{shi2022motion}, without any bells and whistles. This demonstrates that \algname can also serve as a pretraining framework, acting as a foundation model in motion planning to benefit a variety of related tasks.

\begin{table}[h]
\begin{center}
\begin{tabular}{lccccc}
\toprule
\multicolumn{1}{l}{\bf Methods} & scene-centric & mAP $\uparrow$ & minADE $\downarrow$ & minFDE $\downarrow$ & MR $\downarrow$ 
\\ \midrule

MTR-e2e \cite{shi2022motion} & \xmark &\textbf{0.32} & \textbf{0.52} & \textbf{1.10} & \textbf{0.12} \\
CPS~\cite{sun2023large} & \xmark & 0.32 & 0.74 & 1.49 & 0.20 \\
SceneTransformer~\cite{ngiam2021scene} & \cmark   & 0.28 & 0.61 & 1.22 & 0.16 \\
\algname-m & \cmark    & 0.30 & 0.60 & 1.15 & 0.13 \\
\bottomrule
\end{tabular}
\caption{Performance comparison of motion prediction on the validation set of the WOMD. \algname-medium significantly outperforms scene-centric SceneTransformer, and even approaches the performance of agent-centric models like MTR-e2e, without any bells and whistles.}
\label{tab:WODmotion}
\vspace{-8mm}
\end{center}
\end{table}

\subsection{Per-type Results of WOMD Validation}
\label{ape:G2}
We also present the per-type results of \algname on the WOMD Validation set, as shown in reference Tab.~\ref{tab:wod val type}.

\begin{table}[h]
\begin{center}
\begin{tabular}{lccccc}
\toprule
\multicolumn{1}{l}{\bf Object type} & mAP $\uparrow$ & minADE $\downarrow$ & minFDE $\downarrow$ & MR $\downarrow$ 
\\ \midrule
Vehicle & 0.32 & 0.75 & 1.42 & 0.14\\
Pedestrian & 0.27 & 0.36 & 0.70 & 0.07 \\
Cyclist & 0.30 & 0.68 & 1.33 & 0.18 \\
\textbf{Avg} & 0.30 & 0.60 & 1.16 & 0.13 \\
\bottomrule
\end{tabular}
\caption{Per-type performance of motion prediction on the validation set of WOMD.}
\label{tab:wod val type}
\vspace{-8mm}
\end{center}
\end{table}

\section{Per-component WOD Sim Agent Metric}
\label{ape:H}
As shown in Table~\ref{tab:per-metric-scores-test}, to provide a more detailed showcase of our model's performance on the Waymo Sim Agents benchmark, we have listed the breakdown metrics for reference. Our method significantly leads the competition across a majority of these metric components, particularly in linear kinematic metrics, collision-related metrics, and map compliance metrics. It achieves state-of-the-art performance in both the overall realism meta-metric and the minADE (minimum Average Displacement Error) metric, highlighting its effectiveness and efficiency in producing realistic and accurate simulation for the real world driving scenarios.

\begin{table*}[h]
    \centering
    \resizebox{\textwidth}{!}{
    \begin{tabular}{l|cc|ccccccccc}
    \hline
\rowcolor[gray]{0.9} \textbf{\textsc{Agent Policy}}& \cellcolor{Gray}\textbf{\textsc{Meta}}& \textsc{minADE} & \textsc{linear} & \textsc{linear } & \textsc{ang.} & \textsc{ang.} & \textsc{dist.} & \textsc{collision} & \textsc{TTC} & \textsc{dist. to} & \textsc{offroad}  \\
\rowcolor[gray]{0.9} \textsc{} &\cellcolor{Gray}\textbf{\textsc{Metric}}& \textsc{}& \textsc{speed} & \textsc{accel.} & \textsc{speed} & \textsc{accel.} & \textsc{to obj.} & \textsc{} & \textsc{} & \textsc{road edge} & \textsc{}  \\
\rowcolor[gray]{0.9} & \cellcolor{Gray}\textsc{($\uparrow$)} & \textsc{($\downarrow$)}& ($\uparrow$) & ($\uparrow$) & ($\uparrow$) & ($\uparrow$) & ($\uparrow$) & ($\uparrow$) & ($\uparrow$) & ($\uparrow$) & ($\uparrow$)  \\
\hline
\textsc{Logged Oracle}& \cellcolor{Gray}0.722  & 0.000     & 0.561 & 0.330 & 0.563 & 0.489 & 0.485 & 1.000 & 0.881 & 0.713 & 1.000    \\
\textsc{SBTA-ADIA} \cite{Mo23tr_SBTA_ADIA} &  \cellcolor{Gray}0.420& 3.611  & 0.317 & 0.174 & 0.478 & 0.463 & 0.265 & 0.337 & 0.770 & 0.557 & 0.483   \\
\textsc{CAD} \cite{Chiu23tr_CollisionAvoidanceDetour} &  \cellcolor{Gray}0.531& 2.308          & 0.349 & 0.253 & 0.432 & 0.310 & 0.332 & 0.568 & 0.789 & 0.637 & 0.834    \\
\rowcolor[gray]{0.9} \textsc{Joint-Multipath++} \cite{Wang23tr_JointMultipathSimAgents} & \cellcolor{Gray}0.533  & 2.049  & 0.434 & 0.230 & 0.515 & 0.452 & 0.345 & 0.567 & 0.812 & 0.639 & 0.682    \\
\textsc{Wayformer}  \cite{Nayakanti23icra_Wayformer}& \cellcolor{Gray}0.575&  2.498  & 0.331 & 0.098 & 0.413 & 0.406 & 0.297 & 0.870 & 0.782 & 0.592 & 0.866     \\
\rowcolor[gray]{0.9} \textsc{MTR+++} \cite{Qian23tr_SimpleEffectiveSimMultiAgent} &  \cellcolor{Gray}0.608 & 1.679  & 0.414 & 0.107 & 0.484 & 0.436 & 0.347 & 0.861 & 0.797 & 0.654 & 0.895   \\
\textsc{MVTA} \cite{wang2023multiverse} &    \cellcolor{Gray}0.636 &1.866  & 0.439 & 0.220 & 0.533 & 0.480 & 0.374 & 0.875 & 0.829 & 0.654 & 0.893  \\
\rowcolor[gray]{0.9} \textsc{MVTE$^{\star}$} \cite{wang2023multiverse}   &  \textbf{\textbf{\cellcolor{Gray}0.645}}     & 1.674  & 0.445 & 0.222 & 0.535 & 0.481 & 0.383 & 0.893 & 0.832 & 0.664 & 0.908   \\
\textsc{Trajeglish}~\cite{philion2023trajeglish} & \textbf{\cellcolor{Gray}0.644}& 1.615 & 0.448 & 0.192 & \textbf{0.538} & \textbf{0.485} & 0.386 & 0.918 & \textbf{0.837} & 0.658 & 0.887 \\
\textbf{\algname-m}& \textbf{\cellcolor{Gray}0.643}& \textbf{1.590} & \textbf{0.463} & \textbf{0.256} & 0.467 & 0.412 & \textbf{0.391} & \textbf{0.920} & 0.832 & \textbf{0.672} & \textbf{0.908} \\
    \hline
    \end{tabular}
    }
    \caption{Per-component metric results on the \emph{test} split of WOMD, representing likelihoods. Note: $\star$ indicates the use of model ensemble techniques. Our method significantly leads the competition across a majority of metric components and achieves the state-of-the-art on the overall realism meta metric and the minADE metric. The best score is bolded.}
    \label{tab:per-metric-scores-test}
\vspace{-8mm}
\end{table*}

\end{document}